\documentclass[10pt,twocolumn,letterpaper]{article}

\usepackage[pagenumbers]{iccv}      

\usepackage{multirow}
\usepackage{makecell}  

\newcommand{\figref}[1]{Fig.~\ref{#1}}
\newcommand{\tabref}[1]{Tab.~\ref{#1}}
\newcommand{\secref}[1]{Sec.~\ref{#1}}

\usepackage{booktabs}
\usepackage{float}
\usepackage{colortbl}
\usepackage{multirow}
\usepackage{pifont}
\usepackage{overpic}
\usepackage{graphicx}

\usepackage[normalem]{ulem}
\newcommand{\methodname}{Strip R-CNN}
\newcommand{\backbonename}{StripNet}

\newcommand{\myPara}[1]{\vspace{5pt}\noindent\textbf{#1}}

\graphicspath{{images/}}
\renewcommand\arraystretch{1.2} 

\definecolor{iccvblue}{rgb}{0.21,0.49,0.74}
\usepackage[pagebackref,breaklinks,colorlinks,citecolor=iccvblue]{hyperref}

\def\paperID{*****} 
\def\confName{ICCV}
\def\confYear{2025}

\title{\methodname: Large Strip Convolution for Remote Sensing Object Detection}

\author{Xinbin Yuan$^1$ \quad Zhaohui Zheng$^{{1}}$ \quad Yuxuan Li$^1$ \quad Xialei Liu$^1$ \quad Li Liu$^3$ \quad Xiang Li$^{1,2}$ \\ Qibin Hou$^{{1,2}}$\thanks{Corresponding authors.} \quad Ming-Ming Cheng$^{1,2}$\footnotemark[1]  \\ \\
$^1$VCIP, School of Computer Science, NKU \quad $^2$NKIARI, Futian, Shenzhen, China \\ $^3$Academy of Advanced Technology Research of Hunan, Changsha, China \\
 \textit{\normalsize {yxb}@mail.nankai.edu.cn, \{houqb, cmm\}@nankai.edu.cn} \\  
 {\normalsize Project page: \url{https://github.com/HVision-NKU/Strip-R-CNN}}
 }

\begin{document}
\maketitle

\begin{abstract}
While witnessed with rapid development, remote sensing object detection remains challenging for detecting high aspect ratio objects.
This paper shows that large strip convolutions are good feature representation learners for remote sensing object detection and can detect objects of various aspect ratios well.
Based on large strip convolutions, we build a new network architecture called \methodname{}, which is simple, efficient, and powerful.
Unlike recent remote sensing object detectors that leverage large-kernel convolutions with square shapes, our \methodname{} takes advantage of sequential orthogonal large strip convolutions in our  backbone network \textbf{\backbonename{}} to capture spatial information.
In addition, we improve the localization capability of remote-sensing object detectors by decoupling the detection heads and equipping the localization branch with strip convolutions in our \textbf{strip head}.
Extensive experiments on several benchmarks, for example DOTA, FAIR1M, HRSC2016, and DIOR, show that our \methodname{} can greatly improve previous work.
In particular, our 30M model achieves 82.75\% mAP on DOTA-v1.0, setting a new state-of-the-art record.
Our code will be made publicly available.
\end{abstract}

\section{Introduction} \label{sec:intro}

Remote sensing object detection has gained significant attention in recent years due to its application in aerial images captured by drones and satellites~\cite{zaidi_survey_2022,mei2023d2anet,rssurvey2023,li2024predicting,li2024saratr,li2024sardet,long2017accurate,sm3det}.
A popular pipeline is built on the basis of rotated boxes to cover objects of interest.
Due to boundary discontinuity and square-like problems~\cite{yang_rethinking_2021,yang_learning_2021,yang_arbitrary-oriented_2020} and the urgent need to capture long-range information~\cite{lsknet,cai2024poly}, many research breakthroughs have been made to develop stronger rotated object detectors, including object representations~\cite{xie_oriented_2021,xu_gliding_2021,li2022oriented,xiao2024theoretically}, IoU-simulated loss functions~\cite{yang_rethinking_2021,yang_learning_2021,yang_arbitrary-oriented_2020,yang_kfiou_2022}, and foundation models~\cite{lsknet,pu2023adaptive,cai2024poly}, etc.

\begin{figure}[t]
  \centering
  \includegraphics[width=\linewidth]{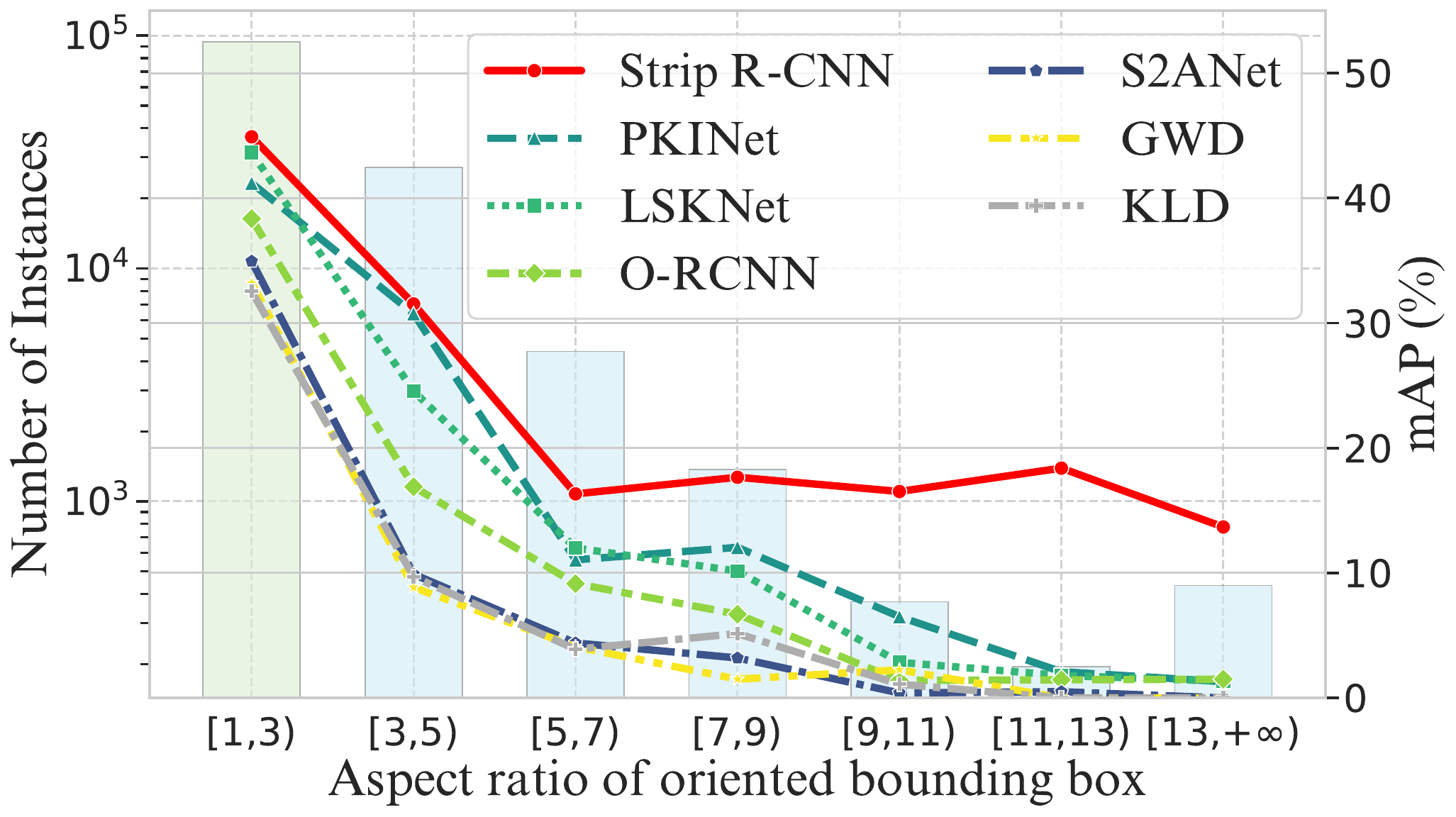}
  \caption{Statistics of the DOTA dataset~\cite{dota_set} and the detection performance of several recent state-of-the-art detectors.
  We can see that slender objects (aspect ratio $>$ 3) occupy a non-negligible proportion and detection performance of previous state-of-the-art models declines as aspect ratio increases. 
  }\label{fig:DotaStatis}
\end{figure}

Despite the great progress made by previous work, successfully detecting high aspect ratio objects, which are prevalent in remote sensing object detection, is still a challenging problem.
To demonstrate this, in~\figref{fig:DotaStatis}, we calculate the statistics of the widely used DOTA dataset~\cite{dota_set}, which shows that slender objects are quite common in remote sensing scenarios and usually occupy a large proportion of the data. 
We also experimentally found that existing object detection methods~\cite{xie_oriented_2021,lsknet,cai2024poly,yang_rethinking_2021,yang_learning_2021,han_align_2020} often struggle with slender objects, where detection performance decreases as the aspect ratio of objects increases. 

We argue that the difficulties in detecting these objects arise from two primary challenges.
First, \emph{high aspect ratio objects contain rich feature information along one spatial dimension, while exhibiting relatively sparse feature in the other.} 
Traditional detectors based on convolutional neural networks mostly extract input feature maps within square windows.
This design greatly restricts their ability to effectively capture the anisotropic context, which can be commonly found in remote sensing images, leading to the inclusion of irrelevant information from surrounding areas.
Second, \emph{high aspect ratio objects pose considerable challenges in regression tasks due to their unique geometric properties.}
In remote sensing object detection, unlike general object detection, an additional angle regression is required. 
For high aspect ratio objects, even a small error in angle estimation can lead to significant deviations from the ground truth.

To date, there are few works considering how to deal with challenging high aspect ratio objects.
A generic approach widely used in previous methods is to enlarge the receptive field of models with large-kernel convolutions~\cite{ding_scaling_2022,ding2024unireplknet,chen2024pelk,liu2022more}.
A typical example should be LSKNet~\cite{lsknet}, which introduces  large-kernel convolutions with a spatial selection mechanism to capture long-range contextual information.
PKINet~\cite{cai2024poly} further extends LSKNet with a parallel large square conv structure to enhance the performance for large variation of object scales.
However, the parallel paradigm of using multiple large-kernel convolutions exacerbates the computational burden, and the complicated block design further restricts model efficiency.
For the high variation of the object aspect ratio, how to efficiently use large-kernel convolutions is still an open question.

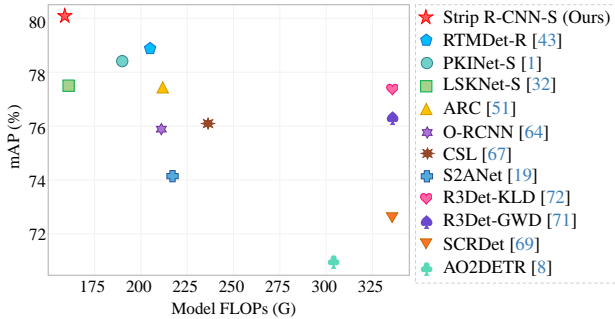
\begin{figure}[t]
  \centering
  \scriptsize
  \begin{overpic}[width=\linewidth]{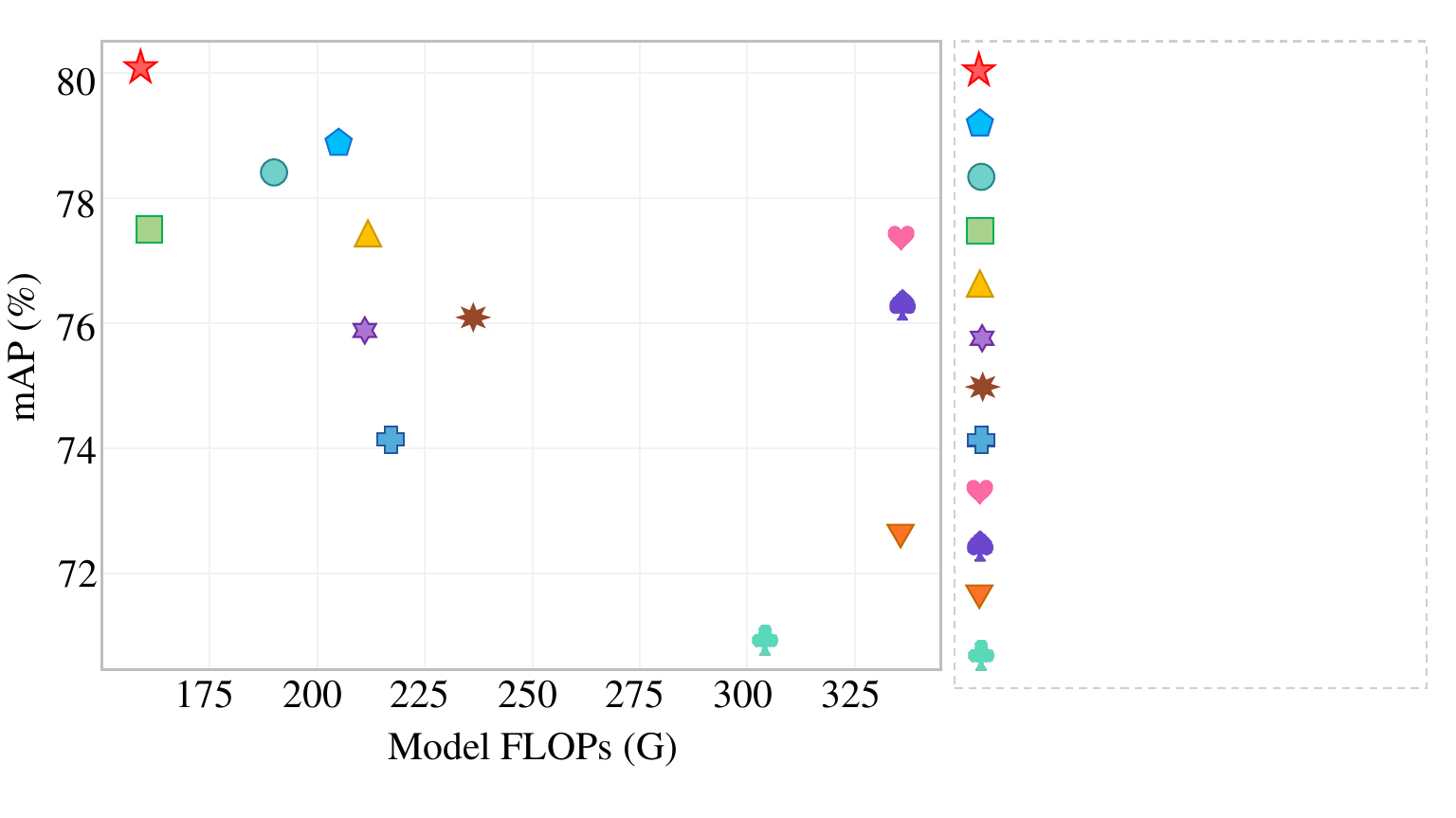}
    \put(70, 50.4){Strip R-CNN-S (Ours)}
    \put(70, 46.8){RTMDet-R~\cite{lyu_rtmdet_2022}}
    \put(70, 43.1){PKINet-S~\cite{cai2024poly}}
    \put(70, 39.7){LSKNet-S~\cite{lsknet}}
    \put(70, 35.8){ARC~\cite{pu2023adaptive}}
    \put(70, 32.1){O-RCNN~\cite{xie_oriented_2021}}
    \put(70, 28.6){CSL~\cite{yang_arbitrary-oriented_2020}}
    \put(70, 25.1){S2ANet~\cite{han_align_2020}}
    \put(70, 21.6){R3Det-KLD~\cite{yang_learning_2021}}
    \put(70, 18.0){R3Det-GWD~\cite{yang_rethinking_2021}}
    \put(70, 14.2){SCRDet~\cite{yang_scrdet_2019}}
    \put(70, 10.5){AO2DETR~\cite{dai_ao2-detr_2022}}
  \end{overpic} \\
  \vspace{-5pt}
  \caption{A comprehensive comparison of detection performance on the DOTA dataset of various remote sensing object detectors. }
  \label{fig:sota}
\end{figure}

In this paper, we propose \methodname{}, which can efficiently combine the advantage of the square convolution and the strip convolution with less feature redundancy.
Our design principles are two-fold. 
First, the new network architecture should be simple and efficient. 
Second, it should be good at handling objects of different aspect ratios even when they are high.
Given the characteristics of the objects in remote sensing images, we propose using orthogonal large strip convolutions as the main spatial filters, which comprise the core component of our \textbf{\backbonename{}} backbone, called strip module.
As such, our network is quite simple but can generalize well to even objects of especially high aspect ratios as shown in \figref{fig:DotaStatis}.
Furthermore, to conquer the second challenge mentioned above, we decouple the detection heads used in previous remote sensing object detections and strengthen the localization branch with strip convolutions in our \textbf{strip head}.
We found that this can not only assist in improving the localization capability of our method but also help more accurately regress the box angles.

To our knowledge, \methodname{} is the first work to explore how to take advantage of large strip convolutions for remote sensing object detection.
Despite simplicity and the lightweight nature, our \methodname{} achieves state-of-the-art performance on the standard DOTA benchmark without bells and whistles, as shown in ~\figref{fig:sota}.
Notably, our model \methodname{-S} with only 30M learnable parameters achieves the best results on DOTA leaderboard with 82.75\% mAP.
We also conduct extensive experiments on several other remote sensing datasets, including FAIR1M, HRSC2016, and DIOR, and show the superiority of our \methodname{} over other methods.
We hope that our design principles could provide new research insights for the remote sensing imagery community.

\section{Related Work}

\myPara{Remote Sensing Object Detection.} 
In remote sensing scenarios, where objects are arbitrarily oriented,
rotated bounding boxes are generally adopted for object representation. 
Early representations of rotated bounding boxes use five parameters \((x,y,w,h,\theta)\)~\cite{ding_learning_2019,yang_scrdet_2019,liu2016ship,yu2023phase}. 
However, due to the periodicity of the angle, training models based on this representation often face boundary discontinuity problems in regression~\cite{yang_arbitrary-oriented_2020,yang_kfiou_2022,yang_rethinking_2021}. 
To address this issue, several approaches propose improved representations~\cite{xie_oriented_2021,xiao2024theoretically,li2022oriented,xu_gliding_2021,wang2019mask,yi2021oriented,fu2020point,yang2021dense} for rotated bounding boxes.
For instance, Oriented R-CNN~\cite{xie_oriented_2021} replaces the angle with midpoint offsets, leading to a six-parameter representation \((x, y, w, h, \triangle\alpha, \triangle\beta)\), which significantly enhances detector performance. 
COBB~\cite{xiao2024theoretically} introduces a continuous representation of rotated boxes with nine parameters based on the aspect ratio of the minimum enclosing rectangle to the rotated bounding box areas.
There are also approaches focusing on mitigating the discontinuity problem in boundary regression through loss functions~\cite{yang_rethinking_2021,yang_learning_2021,hou2023g,cheng_dual-aligned_2022,qian_learning_2021}. 
For example, GWD~\cite{yang_rethinking_2021} and KLD~\cite{yang_learning_2021} convert rotated bounding boxes into 2D Gaussian distributions and use Gaussian Wasserstein Distance and Kullback-Leibler Divergence as the loss functions. 
KFIoU~\cite{yang_kfiou_2022} uses Gaussian modeling and Kalman filtering and propose the SKewIoU Loss. 

Despite the great progress made by previous work, successfully detecting high aspect ratio objects, which are prevalent in remote sensing object detection, is still a challenging problem. 
Our method carefully considers the difficulties posed by these objects and take advantage of large strip convolutions to make our network generalize well to the challenging slender objects, which to our knowledge has not been explored before in this research field.

\begin{figure}[t]
  \centering
  \includegraphics[width=\linewidth]{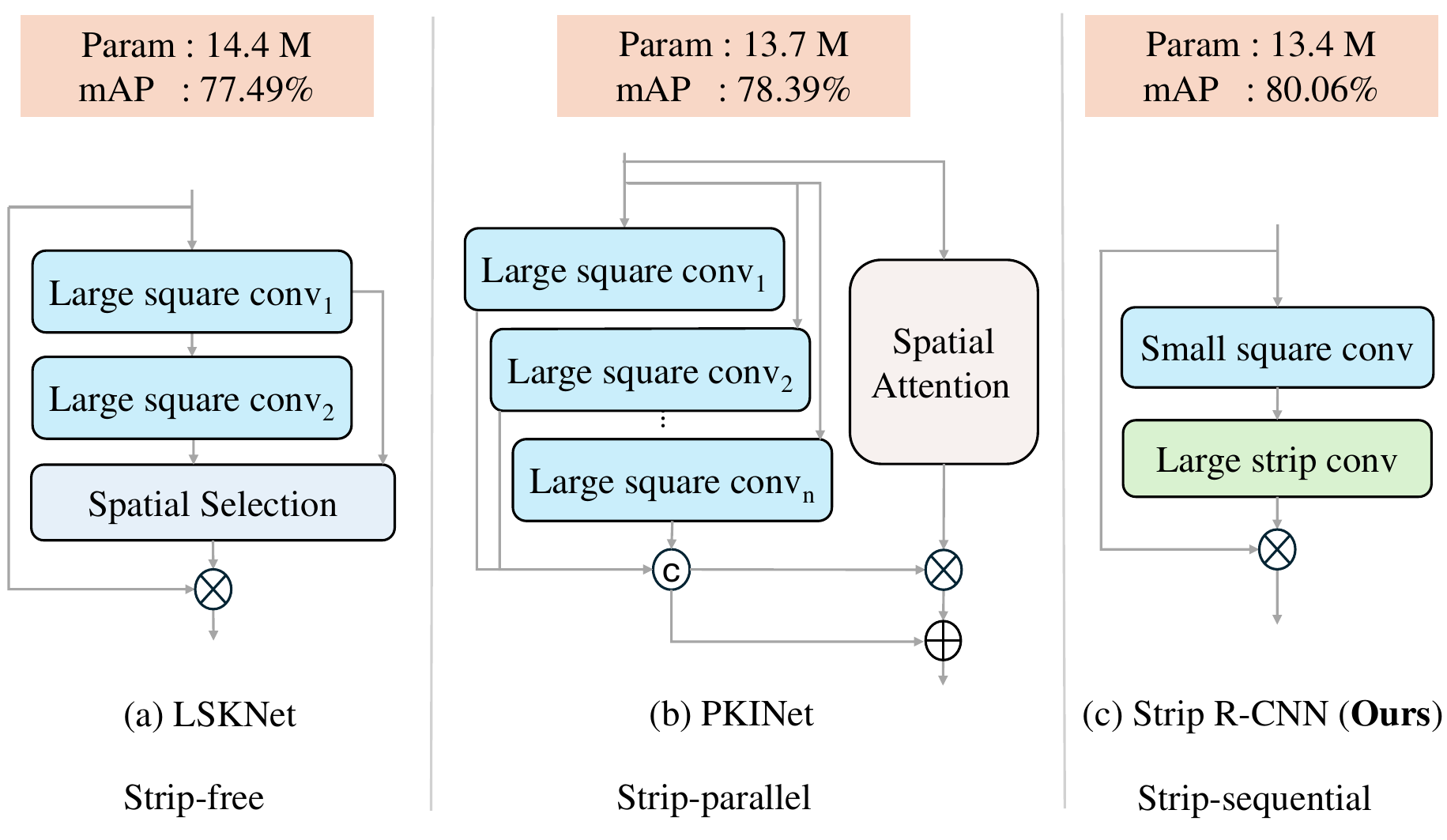}
  \caption{Structural comparison between our proposed strip module and other representative methods using large-kernel convolutions, including LSKNet~\cite{lsknet} and PKINet~\cite{cai2024poly}.
  }\label{fig:comparison}
\end{figure}

\myPara{Large Kernel Networks for Remote Sensing Object Detection.} 
Convolution with large kernel has been an emerging and promising solution to remote sensing object detection, which has been validated to have highly competitive performance against the Transformer-based methods in image classification and segmentation~\cite{hou2024conv2former,guo_visual_2022,guo_segnext_2022,ding_scaling_2022,ding2024unireplknet,chen2024pelk,liu2022more,hou2021coordinate,yin2024dformer}. 
In remote sensing object detection, some approaches put efforts on employing large-kernel convolutions to get long-range contextual information~\cite{Li_2024_IJCV,cai2024poly}. 
For example, LSKNet~\cite{Li_2024_IJCV} utilizes large kernel convolutions and a selection mechanism to model the contextual information needed for different object categories. 
PKINet~\cite{cai2024poly} arranges multiple large-kernel convolutions in parallel to extract dense texture features across diverse receptive fields, and introduces a context anchor attention mechanism to capture relationships between distant pixels. 
However, the parallel paradigm of leveraging multiple large-kernel convolutions exacerbates the computational burden, and the complicated block design makes the model not efficient. 
Regarding the high variation of the object aspect ratio, how to efficiently make use of large-kernel convolutions is still an open question. 
To our knowledge, \methodname{} is the first work to explore how to take advantage of large strip convolutions for remote sensing object detection.

\section{\methodname{}}

In this section, we describe the architecture of the proposed \methodname{} in detail.
Our goal is to advance remote sensing object detectors with large strip convolutions so that the resulting model can perform well on objects of different aspect ratios.
This is different from previous work that emphasizes the importance of convolutions with large square kernels, as shown in \figref{fig:comparison}.

\subsection{Overall Architecture}\label{sec:architecture}

\begin{figure}[t]
  \centering
  \includegraphics[width=0.85\linewidth]{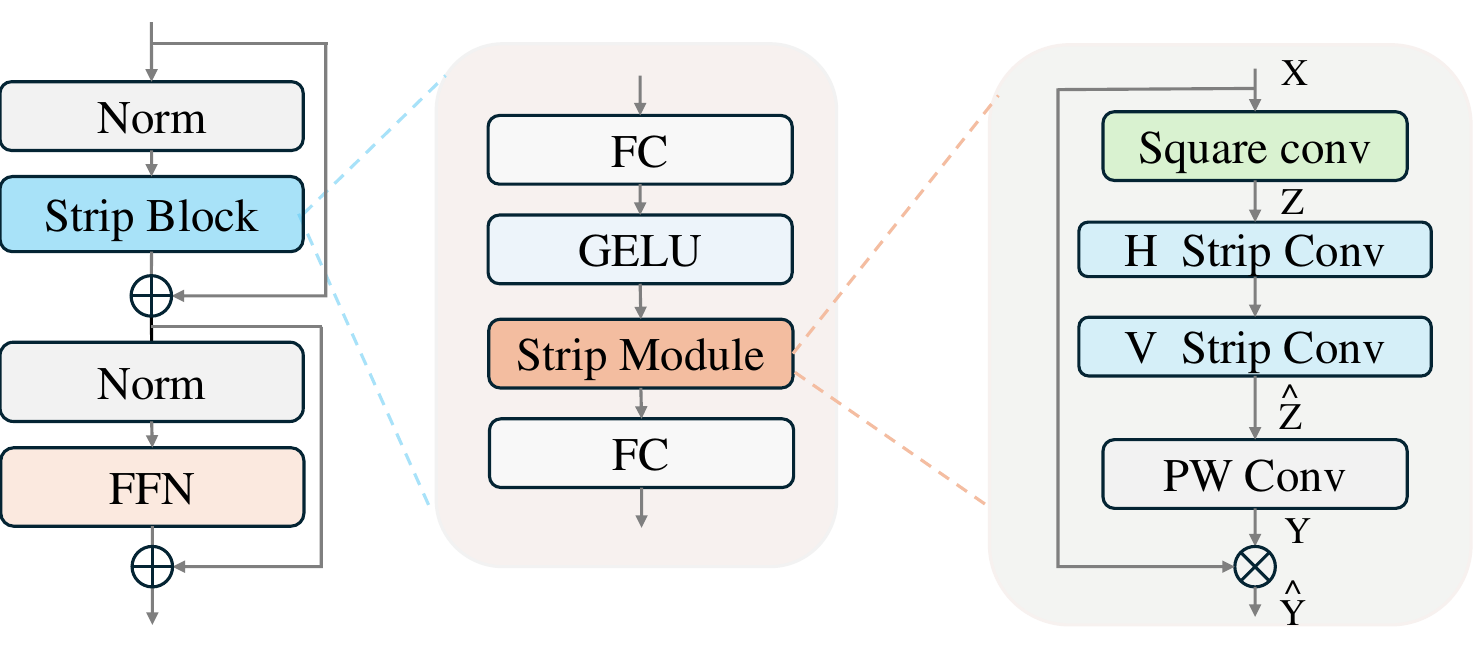}
  \\ \vspace{-5pt}
  \caption{Structure of our basic block of StripNet backbone.}
  \label{fig:strip_module}
\end{figure}

Based on the O-RCNN framework \cite{xie_oriented_2021},
our \methodname{} replace the backbone and detection head with our \textbf{StripNet backbone} and \textbf{strip head}, respectively.
Specifically, the backbone is mainly composed of basic blocks proposed as illustrated in \figref{fig:strip_module}, which consists of two residual sub-blocks: the strip sub-block and the feed-forward network sub-block.
The strip sub-block is built upon a small-kernel standard convolution and two convolutions with large strip-shaped kernels to capture robust features for objects of different aspect ratios.
For the feed-forward network sub-block, we simply follow LSKNet~\cite{lsknet}, which is used to facilitate channel mixing and feature refinement.
The detailed configurations of different variants of our \backbonename{} backbone are shown in \tabref{tab:model_variants}.
For the stem layers, we keep them the same to LSKNet~\cite{lsknet}.
For the detection head, we decouple the localization branch from the original O-RCNN head and enhance it with the proposed strip module, resulting in our strip head, which will be presented in \secref{sec:StripHead}.

\begin{table}[t]
  \centering
  \footnotesize
  \setlength{\tabcolsep}{0.82mm}
  \caption{Variants of \backbonename{} backbone.
    $C_i$: feature channel number; 
    $D_i$: number of strip blocks in each stage $i$.  
  }\label{tab:model_variants}
  \begin{tabular}{llccc} \hline
\textbf{Model} & \{{$C_1$, $C_2$, $C_3$, $C_4$}\} & \{{$D_1$, $D_2$, $D_3$, $D_4$}\} & \#P & FLOPs \\ \hline
$\star$ \backbonename{}-T & \{32, 64, 160, 256\} & \{3, 3, 5, 2\} & 3.8M & 18.2G \\ 
$\star$ \backbonename{}-S & \{64, 128, 320, 512\}& \{2, 2, 4, 2\} & 13.3M & 52.3G \\ \hline 
  \end{tabular} 
\end{table}

\subsection{Strip Module}

As discussed in \secref{sec:intro}, large square kernel convolutions provide essential long-range contextual information for remote sensing applications.
%
PKINet \cite{cai2024poly} introduces parallel large square kernel convolutions and spatial attention mechanism to extract spatial information. However, the parallel paradigm of leveraging multiple large-kernel convolutions exacerbates the computational burden, and the complicated block design makes the model not efficient. 
Our objective is to efficiently extract essential features for objects of varying aspect ratios.
The outcome is a sequential paradigm that efficiently combines the advantages of both standard and strip convolutions without requiring additional information fusion module. 
In what follows, we provide a detailed description of the core of our basic block: \textbf{strip module}.

Given an input tensor \( \mathbf{X} \in \mathbb{R}^{C \times H \times W} \) with \( C \) channels, a depthwise convolution with square kernels \( \mathbf{K} \in \mathbb{R}^{C \times k_H \times k_W }\) is first applied to extract local contextual features, yielding $\mathbf{Z}$, where $H \times W$ and $k_H \times k_W$ are the feature size and kernel size, respectively.
In practical use, we set $k_H \times k_W$ to $5\times5$.
After the initial depthwise convolution, we use two sequential depthwise convolution of large strip-shaped kernels to better capture objects of high aspect ratios.
The output is denoted as $\mathbf{\hat{Z}}$.
Unlike standard convolutions that extract features from a square region for each time, large strip convolutions allow the network to focus more on features along either the horizontal or the vertical axis.
%
The combined use of horizontal and vertical large strip convolutions enables the network to collect directional features across both spatial axes, enhancing the representations of elongated or narrow structures in spatial dimension.

\begin{figure}[t]
  \centering
  \includegraphics[width=0.9\linewidth]{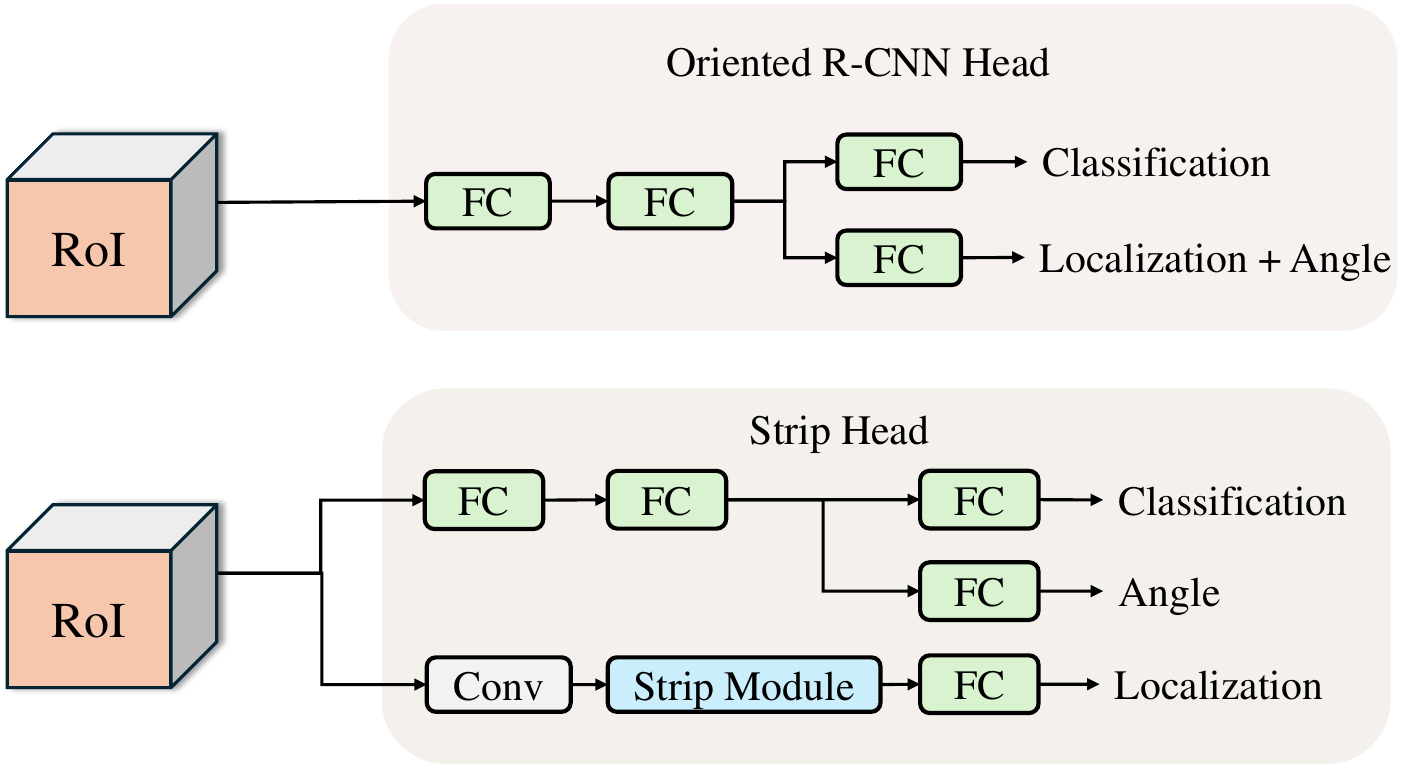}
  \caption{Structural comparison of Oriented R-CNN head and strip head. 
  In our strip head, the classification and angle prediction heads share two fully connected layer, while the localization head incorporates our strip module.
  }\label{fig:strip_head}
\end{figure}

To further enhance the interaction of the features across the channel dimension, a simple point-wise convolution is applied to transform $\mathbf{\hat{Z}}$ to $\mathbf{Y}$.
In this way, each position of the resulting feature map $\mathbf{Y}$ encodes both horizontal and vertical features across a wide spatial area. 
%
%
Finally, following~\cite{guo_segnext_2022,hou2024conv2former}, we regard the feature map $\mathbf{Y}$ as attention weights to reweigh the input $X$, which can be formulated as
\begin{equation}\
\mathbf{\hat{Y}} = \mathbf{X} \cdot \mathbf{Y},
\end{equation}
where `$\cdot$' denotes the element-wise multiplication operation.
\figref{fig:strip_module} provides a diagrammatic illustration of the proposed strip module.
In our experiments, we will discuss how to choose the kernel size of the strip convolutions.

It is important to emphasize that our backbone network \backbonename{} is much simpler than previous remote sensing object detectors using large-kernel convolutions as shown in~\figref{fig:comparison}.
We do not utilize any spatial or channel attention mechanisms in our basic block design nor compound fusion operations with different types of large-kernel convolutions.
This makes our \backbonename{} quite simple but has great performance on different remote sensing detection benchmarks. 
%
Moreover, we found that there is no significant difference in applying horizontal strip convolutions before vertical ones or vice versa.
Both approaches are effective.

\begin{figure}[t]
  \centering
  \includegraphics[width=\linewidth]{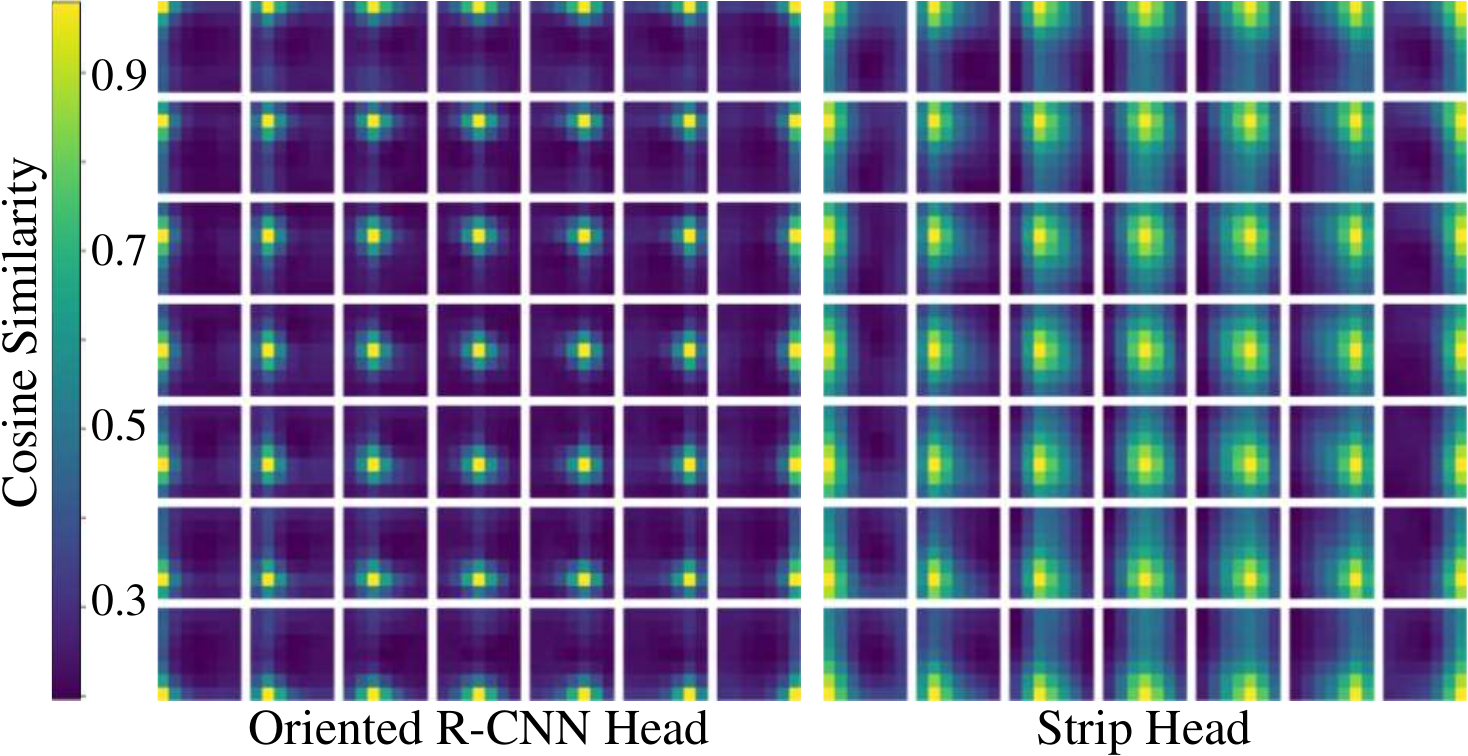}
  \caption{Spatial correlation map comparison of the Oriented R-CNN head and our strip head. 
  Our strip head has significantly more spatial correlations in the output feature maps.
  }\label{fig:correlation_map}
\end{figure}

\begin{figure}[t]
  \centering
  \includegraphics[width=.8\linewidth]{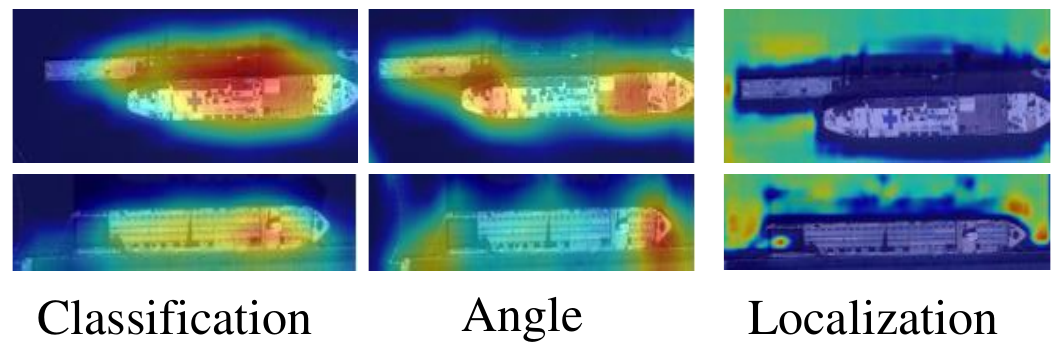}
  \caption{Spatial sensitivity heatmap comparison for classification, localization and angle prediction in the output feature maps.}
    \label{fig:sensitivity_map}
\end{figure}

\subsection{Detection Head with Strip Convolutions}\label{sec:StripHead}

In localization tasks, models should be sensitive to transformations, as the accuracy of localization depends on the positions of input objects. 
Previous strong rotated object detectors~\cite{cai2024poly,lsknet} adopt the Oriented R-CNN framework~\cite{xie_oriented_2021}, whose detection head shares the same linear layers for classification and localization, as shown in~\figref{fig:strip_head}.
However, linear layers have limited spatial correlation as demonstrated in~\cite{wu2020rethinking}, making them transformation-insensitive and unsuitable for precise localization.

A better solution might be to decouple the classification and localization tasks and using small kernel convolutions in the localization branch as suggested in~\cite{wu2020rethinking}.
However, our analysis of spatial correlation maps shows that small kernel convolutions capture only short-range spatial correlations, as illustrated on the left of~\figref{fig:correlation_map}. 
These short-range correlations are inadequate for accurately localizing slender objects, which require long-range dependencies.
To effectively localize objects of varying aspect ratios, we argue that the localization head should be able to capture long-range dependencies, similar to those handled by the backbone network.
%
Large strip convolutions, which capture both horizontal and vertical features across a broad spatial area, provide the extended spatial correlations necessary for better localization. Therefore, we propose using large strip convolutions in the localization branch to enhance the detector’s localization capabilities.

Furthermore, predicting the parameters \(x\), \(y\), \(w\), \(h\), and \(\theta\) together may lead to the issue of coupled features~\cite{yu2024spatial}. To address this, we decouple the prediction of \(\theta\) from the other parameters (\(x\), \(y\), \(w\), \(h\)).
Building on this decoupling approach, we investigate the spatial sensitivity regions of the classification and angle predictions in the output feature maps. The spatial sensitivity heat map is shown in~\figref{fig:sensitivity_map}, where the first column represents the classification sensitivity, and the second shows the angle sensitivity.
Our observation reveals that the spatial sensitivity for classification is concentrated in the central areas of the objects, while the sensitivity for angle is primarily focused near the object borders. The sensitive regions for classification and angle predictions share some overlap and exhibit some complementary characteristics. Based on this, we propose the use of shared fully connected layers for both classification and angle predictions.
The structure of the strip head is shown at the bottom of ~\figref{fig:strip_head}. Detailed settings are described below.


\myPara{Classification head}: 
The classification head simply utilizes two fully connected layers with an output dimension of 1024. 
Using fully connected layers in the classification head has been shown to be a good choice, as demonstrated in Double-Head RCNN~\cite{wu2020rethinking}, so we keep it unchanged.

\myPara{Localization head}: 
The localization head begins with a standard \(3\times3\) convolution to extract local features. 
Then we add a strip module, followed by a fully connected layer to collect long-range spatial dependencies.

\myPara{Angle head}: 
For angle prediction, we adopt three fully connected layers to estimate angular information, which we found works well. 
Note that the first two fully connected layers share parameters with the classification head. 


It is worth noting that our detection head incorporates the proposed strip module.
This design largely improves the detection capability compared to the original Oriented R-CNN head.
As shown in~\figref{fig:correlation_map}, from the visualization of the spatial sensitivity of different methods, we can see that the Oriented R-CNN head has a similar spatial correlation pattern on the output feature maps. 
However, our strip head has much more spatial correlations in the output feature maps.
We will show that this enables our method to be able to better localize objects even of higher aspect ratios.

\section{Experiments}

\subsection{Experiment Setup}

\myPara{Datasets}. We conduct extensive experiments on five popular remote sensing object detection datasets.
\begin{itemize}
\item \textbf{DOTA-v1.0} \cite{dota_set} is a large-scale dataset for remote sensing detection which contains 2,806 images, 188,282 instances, and 15 categories.

\item \textbf{DOTA-v1.5} \cite{dota_set} is a more challenging dataset which contains 403,318 instances and 16 categories.

\item \textbf{FAIR1M-v1.0} \cite{sun_fair1m_2022} is a remote sensing dataset consisting of 15,266 images, 1 million instances, and 37 categories.

\item \textbf{HRSC2016} \cite{hrsc2016} is a remote sensing dataset that contains 1061 aerial images and  
2,976 instances.

\item \textbf{DIOR-R} \cite{DIOR} contains 23,463 images and 192,518 instances.
\end{itemize}

\myPara{Implementation details.}
Our training process is divided into two stages: pretraining on ImageNet~\cite{imagenet} and fine-tuning on downstream remote sensing datasets. For ablation experiments, we train all the models for 100 epochs. To achieve higher performance, we train the models for 300 epochs for the final results. The number of training epochs for DOTA, DOTAv1.5, HRSC2016, FAIR1M-v1.0, and DIOR-R are set to 12, 12, 36, 12, and 12, respectively, following previous methods.
Learning rates are set to 0.0001, 0.0001, 0.0004, 0.0001, and 0.0001, respectively. 
%
%
The input sizes for HRSC2016 and DIOR-R are $800 \times 800$, while for the DOTA-v1.0, DOTA-v1.5 and FAIR1M-v1.0 datasets, the input sizes are $1024 \times 1024$.
During training, we employ the AdamW~\cite{adamw} optimizer with $\beta_1 = 0.9, \beta_2 = 0.999$, and a weight decay of 0.05. All the models are trained on 8 NVIDIA 3090 GPUs with a batch size of 8, and test is conducted on a single NVIDIA 3090 GPU.

\subsection{Main Results}

\begin{table}[t]
  \renewcommand\arraystretch{1.1} 
  \footnotesize
  \setlength{\tabcolsep}{2.5mm}
  \centering
  \caption{Comparisons with different backbone models on DOTA-v1.0. 
    Params and FLOPs are computed for backbone only.
    All backbones are pretrained on ImageNet for 300 epochs. 
    Our \backbonename{}-S achieves higher mAP than previous popular backbones.
  }\label{tab:backbone}
\begin{tabular}{l|c|c|c|c} \hline
  Model (Backbone) & \textbf{\#P$\downarrow$} & \textbf{FLOPs$\downarrow$} & \textbf{FPS} & \textbf{mAP~(\%)} \\ \hline
  ResNet-50~\cite{he2016deep} & 23.3M & 86.1G & 21.8 & 75.87 \\
  LSKNet-S~\cite{lsknet} & 14.4M & 54.4G & 20.7 & 77.49 \\
  PKINet-S~\cite{cai2024poly} & 13.7M & 70.2G & 12.0 & 78.39 \\ \hline
  \rowcolor[rgb]{0.9,0.9,0.9}
  $\star$\backbonename{}-S & 13.3M & 52.3G & 17.7 & 80.06 \\ \hline
\end{tabular}
\end{table}

\begin{table*}[t]
  \renewcommand\arraystretch{1.1} 
  \setlength{\tabcolsep}{1.8pt}
  \footnotesize      
  \centering
  \caption{Comparisons with SOTA methods on the DOTA-v1.0 dataset with single-scale and multi-scale training and testing. 
  The \backbonename{}-S backbone is pretrained on ImageNet for 300 epochs. $^\dagger$: Model ensemble as in MoCAE~\cite{oksuz2023mocae}.
  }\label{tab:dotams}
\begin{tabular}{l|c|cccccccccccccccccc} \hline
  Method &\textbf{Pre.} & \textbf{mAP $\uparrow$} & \textbf{\#P $\downarrow$} & \textbf{FLOPs $\downarrow$} & PL & BD & BR & GTF & SV & LV & SH & TC & BC & ST & SBF & RA & HA & SP & HC \\ \hline
  \multicolumn{20}{c}{\textit{Single-Scale}} \\ \hline
EMO2-DETR~\cite{hu2023emo2} & IN & 70.91 & 74.3M  & 304G 
& 87.99 & 79.46 & 45.74 & 66.64 & 78.90 & 73.90 & 73.30 & 90.40 & 80.55 & 85.89 & 55.19 & 63.62 & 51.83 & 70.15 & 60.04  \\
CenterMap~\cite{wang_learning_2021} & IN   & 71.59   & 41.1M & 198G  &
 89.02 & 80.56 & 49.41 & 61.98 & 77.99 &  74.19 & 83.74 & 89.44 & 78.01 & 83.52 & 47.64 & 65.93 & 63.68 & 67.07 & 61.59 \\
AO2-DETR~\cite{dai_ao2-detr_2022} & IN & 72.15 &  74.3M & 304G 
& 86.01 & 75.92 & 46.02 & 66.65 & 79.70 & 79.93 & 89.17 & 90.44 & 81.19 & 76.00 & 56.91 & 62.45 & 64.22 & 65.80 & 58.96 \\
SASM~\cite{SASM}   & IN   & 74.92 & 36.6M & - 
& 86.42 & 78.97 & 52.47 & 69.84 & 77.30 & 75.99 & 86.72 & 90.89 & 82.63 & 85.66 & 60.13 & 68.25 & 73.98 & 72.22 & 62.37 \\
O-RCNN~\cite{xie_oriented_2021}    & IN   & 75.87   & 41.1M &  199G & 
89.46 & 82.12 & 54.78 & 70.86 & 78.93 & 83.00 & 88.20 & 90.90 & 87.50 & 84.68 & 63.97 & 67.69 & 74.94 & 68.84 & 52.28 \\
R3Det-GWD~\cite{yang_rethinking_2021}   & IN    & 76.34 & 41.9M & 336G & 
88.82 & 82.94 & 55.63 & 72.75 & 78.52 &  83.10 & 87.46 & 90.21 & 86.36 & 85.44 & 64.70 & 61.41 & 73.46 & 76.94 & 57.38 \\
COBB~\cite{xiao2024theoretically} & IN & 76.52 & 41.9M & - &
- & - & - & - & - & - & - & - & - & - & - & - & - & - & - \\
R3Det-KLD~\cite{yang_learning_2021}  & IN   & 77.36   & 41.9M & 336G &
88.90 & 84.17 & 55.80 & 69.35 & 78.72 & 84.08 & 87.00 & 89.75 & 84.32 & 85.73 & 64.74 & 61.80 & 76.62 & 78.49 & 70.89 \\
ARC~\cite{yang_kfiou_2022}  & IN   & 77.35   & 74.4M & 217G  & 
89.40 & 82.48 & 55.33 & 73.88 & 79.37 & 84.05 & 88.06 & 90.90 & 86.44 & 84.83 & 63.63 & 70.32 & 74.29 & 71.91 & 65.43 \\
LSKNet-S~\cite{lsknet} & IN &   77.49   & 31.0M  & 161G  & 
89.66 & 85.52 & 57.72 & 75.70 & 74.95 & 78.69 & 88.24 & 90.88 & 86.79 & 86.38 & 66.92 & 63.77 & 77.77 & 74.47 & 64.82  \\
PKINet-S~\cite{cai2024poly} & IN & 78.39 & 30.8M & 190G  &
89.72 & 84.20 & 55.81 & 77.63 & 80.25 & 84.45 & 88.12 & 90.88 & 87.57 & 86.07 & 66.86 & 70.23 & 77.47 & 73.62 & 62.94 \\
RTMDet-R~\cite{lyu_rtmdet_2022} & IN & 78.85 & 52.3M & 205G & 89.43 & 84.21 & 55.20 & 75.06 & 80.81 & 84.53 & 88.97 & 90.90 & 87.38 & 87.25 & 63.09 & 67.87 & 78.09 & 80.78 & 69.13 \\
\hline
\rowcolor[rgb]{0.9,0.9,0.9}$\star$ \methodname{}-S  & IN &  \textbf{80.06}  & \textbf{30.5M}  & \textbf{159G}  & 
88.91 &  86.38 & 57.44 &  76.37 &  79.73 &  84.38 &  88.25 &  90.86 &  86.71 &  87.45 &  69.89 &  66.82 &  79.25 &  82.91 &  75.58 \\
\hline
\multicolumn{20}{c}{\textit{Multi-Scale}}  \\ 
\hline
CSL~\cite{yang_arbitrary-oriented_2020}   & IN    & 76.17   & 37.4M & 236G  & 90.25 & 85.53 & 54.64 & 75.31 & 70.44 & 73.51 & 77.62 & 90.84 & 86.15 & 86.69 & 69.60 & 68.04 & 73.83 & 71.10 & 68.93  \\
DODet~\cite{DODet}   & IN  & 80.62   &  -&  - & 89.96 & 85.52 & 58.01 & 81.22 & 78.71 & 85.46 & 88.59 & {90.89} & 87.12 & 87.80 & 70.50 & 71.54 & 82.06 & 77.43 & 74.47  \\
AOPG~\cite{DIOR} & IN   & 80.66 &-& - & 89.88 & 85.57 & 60.90 & 81.51 & 78.70 & 85.29 & 88.85 & {90.89} & 87.60 & 87.65 & 71.66 & 68.69 & 82.31 & 77.32 & 73.10  \\
KFloU~\cite{yang_kfiou_2022}  & IN   & 80.93   & 58.8M & 206G  & 89.44 & 84.41 & 62.22 & 82.51 & 80.10 & 86.07 & 88.68 & 90.90 & 87.32 & 88.38 & 72.80  & 71.95 & 78.96 & 74.95 & 75.27  \\ 
PKINet-S~\cite{cai2024poly} & IN & 81.06 & 30.8M & 190G  &
89.02 & 86.73 & 58.95 & 81.20 & 80.41 & 84.94 & 88.10 & 90.88 & 86.60 & 87.28 & 67.10 & 74.81 & 78.18 & 81.91 & 70.62 \\
RTMDet-R~\cite{lyu_rtmdet_2022}   & CO    & 81.33  & 52.3M & 205G & 88.01 & 86.17 & 58.54 & 82.44 & 81.30 & 84.82 & 88.71 & {90.89} & 88.77 & 87.37 & 71.96 & 71.18 & 81.23 & 81.40 & 77.13  \\ 
RVSA~\cite{wang_advancing_2022}  & MA & 81.24   & 114.4M & 414G & 88.97 & 85.76 & 61.46 & 81.27 & 79.98 & 85.31 & 88.30  & 90.84 & 85.06 & 87.50 & 66.77 & 73.11 & 84.75 & 81.88 & 77.58  \\ 
LSKNet-S~\cite{lsknet} & IN &   81.64   & 31.0M  & 161G  &  89.57   & 86.34 &  63.13 & 83.67  & 82.20  &  86.10 & 88.66 & {90.89}  & 88.41  & 87.42 & 71.72 & 69.58 & 78.88  & 81.77 & 76.52  \\
\hline
\rowcolor[rgb]{0.9,0.9,0.9}$\star$ \methodname{}-T  & IN &    81.40    & 20.5M &  123G  &   89.14    &  84.90     & 61.78      & 83.50     & 81.54      & 85.87      & 88.64      & {90.89}      & 88.02      & 87.31      & 71.55      & 70.74      & 78.66      & 79.81      & 78.16       \\
\rowcolor[rgb]{0.9,0.9,0.9}$\star$ \methodname{}-S  & IN &  \textbf{82.28}  & \textbf{30.5M}  & \textbf{159G}  &  89.17   &  85.57 &   62.40 & 83.71  & 81.93  &  86.58 & 88.84 & 90.86 & 87.97  & 87.91 & 72.07 & 71.88 & 79.25  & 82.45 & 82.82  \\ 
\rowcolor[rgb]{0.9,0.9,0.9}$\star$ \methodname{}-S$^\dagger$  & IN &  \textcolor{RoyalBlue}{\textbf{82.75}}   & 30.5M  & 159G  &  88.99   &  86.56 &   61.35 & 83.94  & 81.70  &  85.16 & 88.57 & 90.88  & 88.62  & 87.36 & 75.13 & 74.34 & 84.58  & 81.49 & 82.56 \\ \hline
\end{tabular}
\end{table*}

\begin{table*}[t]
\renewcommand\arraystretch{1.2} 
\footnotesize      
\centering
\caption{Comparisons with SOTA methods on the DOTA-v1.5 dataset with single-scale training and testing. The \backbonename{}-S backbone is pretrained on ImageNet for 300 epochs.}
\label{tab:dota15ss}
\setlength{\tabcolsep}{3.2pt}
\begin{tabular}{l|c|c|c|c|c|c|c|c|c} \hline
Method     & RetinaNet-O ~\cite{retinanet}& FR-O ~\cite{frcnn} &  Mask RCNN~\cite{maskrcnn} & HTC~\cite{Chen2019HybridTC} & ReDet ~\cite{han_redet_2021}& DCFL~\cite{xu2023dynamic} & LSKNet-S~\cite{lsknet} & PKINet-S ~\cite{cai2024poly}& \cellcolor[rgb]{0.9,0.9,0.9}\textbf{\methodname{}-S} \\ 
\hline
\textbf{mAP~(\%)} & 59.16 & 62.00 & 62.67 & 63.40 & 66.86 & 67.37 & 70.26 & 71.47 & \cellcolor[rgb]{0.9,0.9,0.9} \textbf{72.27}\\ \hline
\end{tabular}
\end{table*}

\begin{table*}[t]
\renewcommand\arraystretch{1.2} 
\footnotesize 
\centering 
\setlength{\abovecaptionskip}{3pt}
\caption{Comparisons with SOTA methods on the FAIR1M-v1.0 dataset. The \backbonename{}-S backbone is pretrained on ImageNet for 300 epochs. *: Results are referenced from the FAIR1M paper~\cite{sun_fair1m_2022}.}
\label{tab:fair1m}
\setlength{\tabcolsep}{4.4pt}
\begin{tabular}{l|c|c|c|c|c|c|c|c}  \hline
Method  & G. V.*~\cite{xu_gliding_2021} & RetinaNet*~\cite{retinanet} & C-RCNN*~\cite{cascade_rcnn} & F-RCNN*~\cite{frcnn}  & RoI Trans.*~\cite{ding_learning_2019}  & O-RCNN~\cite{xie_oriented_2021}   &  LSKNet-S~\cite{lsknet} &\cellcolor[rgb]{0.9,0.9,0.9} \textbf {\methodname{}-S} \\ 
\hline
\textbf{mAP (\%)}     & 29.92    & 30.67    & 31.18  & 32.12  & 35.29  & 45.60     & 47.87 & \cellcolor[rgb]{0.9,0.9,0.9} \textbf{48.26}  \\ \hline
\end{tabular}
\end{table*}

We first compare our \methodname{} with recent state-of-the-art methods with strong backbones implemented within the Oriented R-CNN~\cite{xie_oriented_2021} framework on the DOTA v1.0 dataset.
As shown in~\tabref{tab:backbone}, \methodname{-S} achieves an improvement of 1.67\% while using 0.4\% fewer parameters and only 74.3\% of the computations required by PKINet-S~\cite{cai2024poly}. Additionally, \methodname{-S} shows a 2.57\% enhancement over LSKNet-S~\cite{lsknet} utilizing 1.1\% fewer parameters and 2.2\% less computations.

\myPara{Results on DOTA-v1.0~\cite{dota_set}}. We conduct a comparative analysis of different models and present detailed results for mean Average Precision (mAP) and Average Precision (AP) across categories on the DOTA dataset(refer to more model results in Supplementary Materials). 
As shown in~\tabref{tab:dotams}, our single-scale evaluation demonstrates a 1.67\% improvement over PKINet-S. Furthermore, with multi-scale training and testing, we achieve 82.28\% mAP for a single model. 
By ensembling the results of RTMDet and \methodname{}, following the model ensemble strategy in MoCAE~\cite{oksuz2023mocae}, we achieve 82.75\% mAP, setting a new state-of-the-art record.

\begin{table}[t]
  \renewcommand\arraystretch{1.1} 
  \centering
  \footnotesize
  \setlength{\tabcolsep}{1mm}
  \caption{Comparisons with SOTA methods on HRSC2016. 
    mAP (07/12): VOC 2007~\cite{voc2007}/2012~\cite{voc2012} metrics. 
    The \backbonename{}-S backbone is pretrained on ImageNet for 300 epochs.
  }
  \label{tab:hrsc}
\begin{tabular}{l|c|cccc}  \hline
Method & \small\textbf{Pre.} & \small\textbf{mAP(07)$\uparrow$} & \small\textbf{mAP(12)$\uparrow$}  & \small\textbf{\#P $\downarrow$} & \small\textbf{FLOPs $\downarrow$}\\ 
\hline
DRN~\cite{pan_dynamic_2020} & IN & - & 92.70 & - &  - \\
DAL~\cite{ming_dynamic_2021}                    & IN     & 89.77                       & -       &  36.4M & 216G    \\
GWD~\cite{yang_rethinking_2021}                    & IN      & 89.85                       & 97.37       &  47.4M & 456G  \\
AOPG~\cite{DIOR}                   & IN      & 90.34                      & 96.22        &  - &  -  \\
O-RCNN~\cite{xie_oriented_2021}         & IN    & 90.50                        & 97.60               &  41.1M  & 199G  \\
RTMDet~\cite{lyu_rtmdet_2022}                 & CO     & 90.60                       & 97.10        & 52.3M  & 205G  \\
LSKNet-S~\cite{lsknet}     & IN     &     90.65                           &       98.46        & 31.0M  & 161G   \\
PKINet-S~\cite{cai2024poly}    & IN     &   90.65                            &        98.54         & 30.8M  & 190G   \\
\hline
\rowcolor[rgb]{0.9,0.9,0.9}$\star$ \methodname{}-S     & IN     &      90.60                           &        \textbf{98.70}         & 30.5M & 159G   \\ \hline
\end{tabular}
\end{table}

\begin{table}[htbp]
\renewcommand\arraystretch{1.1} 
\footnotesize      
\centering
\caption{Comparisons with SOTA methods on DIOR-R. The \backbonename{}-S backbone is pretrained on ImageNet for 300 epochs.}
\label{tab:dior}
\setlength{\tabcolsep}{2.3mm}
\begin{tabular}{l|c|c|c|c} \hline
Method& \textbf{Pre.} & \small\textbf{\#P $\downarrow$} & \small\textbf{FLOPs $\downarrow$} & \textbf{mAP~(\%)} \\
\hline

RetinaNet-O~\cite{retinanet} & IN & - & - & 57.55 \\
Faster RCNN-O~\cite{frcnn} & IN & 41.1M & 198G  & 59.54 \\
TIOE-Det~\cite{ming2023task}  & IN &41.1M & 198G & 61.98 \\
O-RCNN~\cite{xie_oriented_2021} & IN & 41.1M & 199G &64.30 \\
ARS-DETR~\cite{zeng2024ars} & IN & 41.1M & 198G & 66.12 \\
O-RepPoints~\cite{li2022oriented} & IN & 36.6M & - & 66.71 \\
DCFL~\cite{xu2023dynamic} & IN &  - &  - & 66.80 \\
LSKNet-S~\cite{lsknet} & IN &  31M & 161G  & 65.90 \\
PKINet-S~\cite{cai2024poly} & IN &  30.8M & 190G  & 67.03 \\
\hline
\rowcolor[rgb]{0.9,0.9,0.9}
$\star$ \methodname{}-S & IN & 30.5M & 159G & \textbf{68.70} \\ \hline

\end{tabular}
\end{table}

\myPara{Results on DOTA-v1.5~\cite{dota_set}}. In this dataset with minuscule instances, as shown in~\tabref{tab:dota15ss}, our approach achieves outstanding performance, demonstrating its efficacy and generalization ability to small objects. Our \methodname{} outperforms the former state-of-the-art method, achieving an improvement of 0.8\%.

\myPara{Results on FAIR1M-v1.0~\cite{sun_fair1m_2022}}. The results in~\tabref{tab:fair1m} reveal that our \methodname{} reaches a highly competitive mAP score of 48.26\%. Our method could improve 0.39\% mAP for LSKNet~\cite{lsknet} and surpassing all other methods by a significant margin.

\myPara{Results on DIOR-R~\cite{DIOR}}. As shown in~\tabref{tab:dior}, we observe 2.80\% improvement over LSKNet~\cite{lsknet} and 1.67\% improvement over PKINet~\cite{cai2024poly}. 

\myPara{Results on HRSC2016~\cite{hrsc2016}}. We achieve 98.70\% mAP under the VOC2012~\cite{voc2012} metric, which is the state-of-the-art performance with 0.16\% mAP improvement over PKINet~\cite{cai2024poly} and 0.24\% mAP over LSKNet~\cite{lsknet}. The results are shown in ~\tabref{tab:hrsc}.

Across multiple datasets, our method consistently surpasses previous state-of-the-art approaches, demonstrating its generalizability and effectiveness.

\subsection{Ablation Studies}

\begin{figure}[t]

    \centering
    \includegraphics[width=\linewidth]{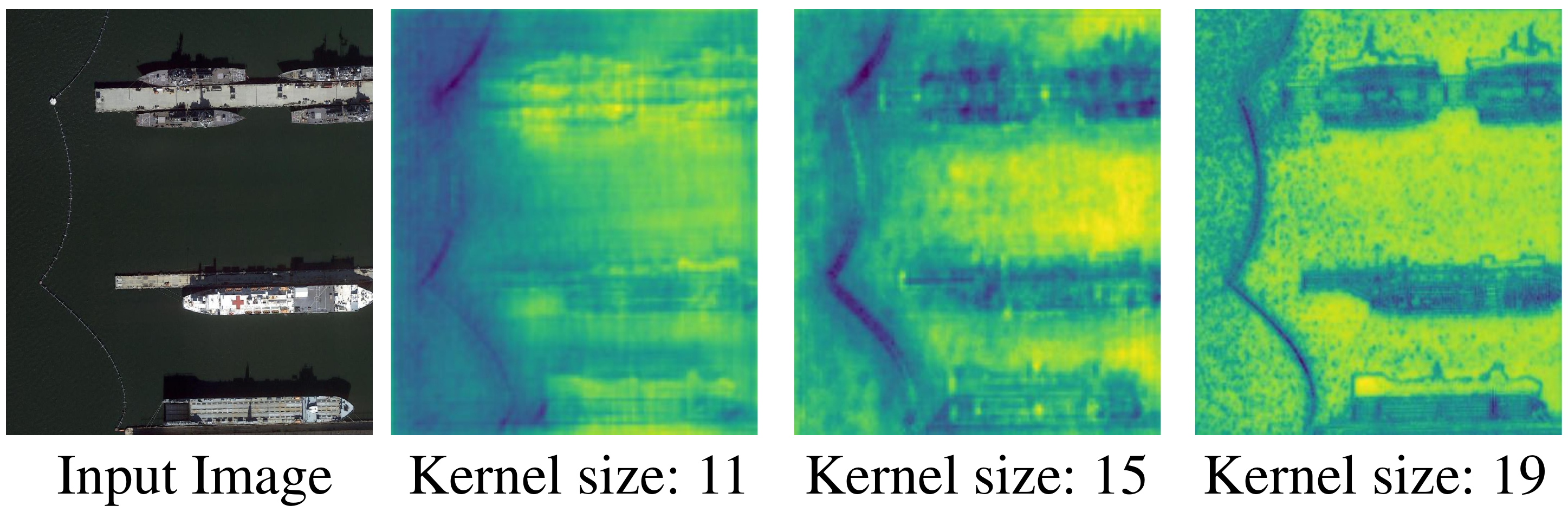}
    \caption{Feature map visulization of different kernel sizes. Kernel size 19 leads to more accurate features of high aspect ratio objects.}
    \label{fig:kernel_size}

\end{figure}

\textbf{Kernel size of strip convolutions.} 
The kernel size in strip convolutions is critical for our proposed strip module. We experiment with large kernel sizes, starting from 11 and increasing in increments of 4. Smaller kernels are ineffective in capturing features of high aspect ratio objects, while a kernel size of 15 yielded satisfactory results, with further improvement observed at 19. To further show the advantage of large strip convolution, we add a visual analysis of the feature representations of different kernel sizes.
\figref{fig:kernel_size} shows that strip convs with a larger kernel size can learn the features of slender objects with more precise localization information and well sharpen the object boundaries.
We also test various kernel sizes at four stages in our \backbonename{} backbone, exploring both increasing and decreasing strategies. Larger kernels in shallow layers combined with smaller kernels in deeper layers produce good results. In contrast, using smaller kernels in the shallow layers leads to significant performance drops, indicating that shallow layers benefit from larger receptive fields. Consequently, we select 19 as the optimal kernel size for the strip module at all stages of StripNet backbone.

\begin{table}[t]
\renewcommand\arraystretch{1.1} 
\small
\setlength{\tabcolsep}{3.9mm}
\centering
\caption{Ablation study on the kernel size of our proposed strip module at four stages of the \backbonename{} backbone network. We adopt the \backbonename{}-S backbone pretrained on ImageNet for 100 epochs. The best result is obtained when using kernel size 19 at all stages.}
\label{tab:kernel_size}
\begin{tabular}{c|c|c|c}  \hline
Kernel Size & \small\textbf{\#P$\downarrow$} & \small\textbf{FLOPs$\downarrow$} & \textbf{mAP}~(\%) \\ 
\hline
\rowcolor[rgb]{0.9,0.9,0.9}
(19,19,19,19)      &  13.30M  & 52.34G & \textbf{81.75}                \\
(15,15,15,15)      &  13.28M  & 52.19G & 81.64            \\
(11,11,11,11)      &   13.26M & 52.03G & 81.22            \\
(15,17,19,21)      &  13.31M  & 52.26G & 81.37              \\
(21,19,17,15)      &  13.29M  & 52.34G  &  81.72  \\  \hline

\end{tabular}
\end{table}

\begin{table}[t]
\renewcommand\arraystretch{1.2} 
\footnotesize
\setlength{\tabcolsep}{1.3mm}
\centering
\caption{Ablation study on the design of our proposed strip module. We adopt the \backbonename{}-S backbone pretrained on ImageNet for 100 epochs. d: dilation rate.}
\label{tab:strip module}
\begin{tabular}{c|c|c|c|c|c} \hline
{$5 \times 5$}& \multicolumn{2}{c|}{Large Strip Conv}  & \multirow{2}{*}{\small\textbf{\#P $\downarrow$}} & \multirow{2}{*} 
 {\small\textbf{FLOPs $\downarrow$}} & \multirow{2}{*} {\textbf{mAP}~(\%)}   \\ 
\cline{2-3}
Square Conv & Sequential & Parallel                  &      &         &   \\ 
\hline
\ding{55} & \ding{51}     &  \ding{55}         &  13.23M  &  51.84G  &  81.38           \\
 \rowcolor[rgb]{0.9,0.9,0.9}
  \ding{51}          &  \ding{51}      &    \ding{55}       &   13.30M  & 52.34G  &    \textbf{81.75}           \\

\ding{51} & \ding{55}   &     \ding{51}         &  13.33M & 52.52G  &  81.54 \\
\hline

\multirow{2}{*}{\ding{51}} & \multicolumn{2}{c|}{\(19\times19\)} & \multirow{2}{*}{14.17M} & \multirow{2}{*}{58.41G} & \multirow{2}{*}{81.44} \\
& \multicolumn{2}{c|}{Square Conv}&  & &\\
\hline
\multirow{2}{*}{\ding{51}} & \multicolumn{2}{c|}{\(7\times7\)} & \multirow{2}{*}{13.30M} & \multirow{2}{*}{52.34G} & \multirow{2}{*}{81.55} \\
& \multicolumn{2}{c|}{Square Conv d=3}&  & &\\ \hline
\end{tabular}
\end{table}

\myPara{Ablation on strip module design.}
We perform ablation experiments to analyze the design choices in the strip module as shown in~\tabref{tab:strip module}. First, we assess the role of depth-wise square convolution. Removing this component leads to a large performance drop, emphasizing the importance of square convolution for capturing features of square-shaped objects.
Next, we examine the integration of horizontal and vertical large strip convolutions, comparing parallel and sequential arrangements. The sequential configuration outperforms the parallel one, as the latter lacks effective two-dimensional modeling, merely combining the two strip convolutions without capturing the overall object structure.
Furthermore, substituting the sequential large strip convolutions with either a \(19 \times 19\) large kernel convolution or a \(7 \times 7\) dilated convolution with a dilation rate of 3 results in noticeable performance loss, further validating the effectiveness of the large strip convolutions.

\begin{table}[t]
\renewcommand\arraystretch{1.2} 
\footnotesize      
\centering
\caption{Effectiveness of \backbonename{}-S backbone on other remote sensing object detection frameworks. The \backbonename{}-S backbone is pretrained on ImageNet~\cite{imagenet} for 100 epochs.}
\label{tab:stripnet_backbone_sup}
\setlength{\tabcolsep}{2.7pt}
\begin{tabular}{l|c|ccc} 
\hline
Method                              &\textbf{Backbone}  &\small\textbf{\#P$\downarrow$} & \small\textbf{FLOPs$\downarrow$} &\textbf{mAP}~(\%)   \\ 
\hline
\multirow{2}{*}{RoI Trans.~\cite{ding_learning_2019}}    & ResNet-50~\cite{he2016deep} & 41.1M & 211.4G &74.61 
 \\
&  \cellcolor[rgb]{0.9,0.9,0.9}\backbonename{}-S &\cellcolor[rgb]{0.9,0.9,0.9}\textbf{13.3M} &\cellcolor[rgb]{0.9,0.9,0.9}\textbf{52.3G}&\cellcolor[rgb]{0.9,0.9,0.9}\textbf{81.72}  \\

\hline

\multirow{2}{*}{S$^2$ANet~\cite{han_align_2020}}     & ResNet-50~\cite{he2016deep} & 41.1M & 211.4G &79.42 
 \\
&  \cellcolor[rgb]{0.9,0.9,0.9}\backbonename{}-S &\cellcolor[rgb]{0.9,0.9,0.9}\textbf{13.3M} &\cellcolor[rgb]{0.9,0.9,0.9}\textbf{52.3G}&\cellcolor[rgb]{0.9,0.9,0.9}\textbf{80.52}  \\

\hline

\multirow{2}{*}{R3Det~\cite{yang_r3det_nodate}}    & ResNet-50~\cite{he2016deep} & 41.1M & 211.4G &76.47 
 \\
&  \cellcolor[rgb]{0.9,0.9,0.9}\backbonename{}-S &\cellcolor[rgb]{0.9,0.9,0.9}\textbf{13.3M} &\cellcolor[rgb]{0.9,0.9,0.9}\textbf{52.3G}&\cellcolor[rgb]{0.9,0.9,0.9}\textbf{79.01}  \\
\hline

\multirow{2}{*}{YOLO~\cite{lyu_rtmdet_2022}}    & CSPNeXt-l~\cite{he2016deep} & 52.3M & 80.2G & 78.02 
 \\
&  \cellcolor[rgb]{0.9,0.9,0.9}\backbonename{}-S &\cellcolor[rgb]{0.9,0.9,0.9}\textbf{13.3M} &\cellcolor[rgb]{0.9,0.9,0.9}\textbf{52.3G}&\cellcolor[rgb]{0.9,0.9,0.9}\textbf{78.50}  \\
\hline
\multirow{2}{*}{ARS-DETR~\cite{zeng2024ars}}    & ResNet-50~\cite{he2016deep} & 41.1M & 211.4G &72.08 
 \\
&  \cellcolor[rgb]{0.9,0.9,0.9}\backbonename{}-S &\cellcolor[rgb]{0.9,0.9,0.9}\textbf{13.3M} &\cellcolor[rgb]{0.9,0.9,0.9}\textbf{52.3G}&\cellcolor[rgb]{0.9,0.9,0.9}\textbf{75.07}  \\
\hline

\end{tabular}
\end{table}

\myPara{Effectiveness of the \backbonename{} backbone.}
To validate the generality and effectiveness of our proposed \backbonename{} backbone, we conduct experiments using various remote sensing detection frameworks. 
%
%
As can be seen in~\tabref{tab:stripnet_backbone_sup},
%
Our method improves RoI Transformer~\cite{ding_learning_2019} by 7.11\%. It also enhances S$^2$ANet and R3Det by 0.9\% and 2.54\%, respectively. When integrated into YOLO and DETR frameworks, our approach consistently boosts performance, demonstrating its broad compatibility and effectiveness in enhancing various detectors.

\myPara{Ablation on strip head design.}
In designing the strip head, we initially decouple the head into \((x, y)\), \((w, h)\), and \(\theta\) branches, applying the strip module to either the \((x, y)\) or \((w, h)\) branches. This approach shows moderate improvement. Combining the \((x, y)\) and \((w, h)\) branches yields slightly better results, probably due to the complementary and similar features shared by  object position and shape information. 
For the \(\theta\) branch, we found that fully connected layers slightly outperform convolutional layers. Therefore, we adopt two fully connected layers for the \(\theta\) branch, and share them with the classification branch in that the sensitive regions for classification and angle predictions share some overlap and exhibit some complementary characteristics, as discussed in~\secref{sec:StripHead}. Simply applying the strip module to the localization branch produces noticeable improvements. We argue that more specialized structures tailored to different regression parameters could further enhance performance. Results are shown in~\tabref{tab:strip head design}.
\begin{table}[t]
\renewcommand\arraystretch{1.1} 
\footnotesize
\setlength{\tabcolsep}{6mm}
\centering
\caption{Ablation study of the \textbf{strip head design}. \textit{fc} refers to using two fully connected layers. \textit{conv} refers to using two $3\times3$ convolutional layers. \textit{strip} refers to replacing the latter $3\times3$ convolutional layer with our designed strip module. We adopt the \backbonename{}-S backbone pretrained on ImageNet for 100 epochs.}
\label{tab:strip head design}
\begin{tabular}{c|c}  \hline
Different Head Structure & \textbf{mAP}~(\%) \\ 
\cline{1-2}
\((x,y,w,h,cls,\theta)_{fc}\)        &  81.75                        \\
\((x,y)_{conv},(w,h)_{conv},(cls)_{fc},(\theta)_{fc}\) & 81.76       \\
\((x,y)_{conv},(w,h)_{conv},(cls)_{fc},(\theta)_{conv}\) & 81.72       \\
\((x,y)_{strip},(w,h)_{conv},(cls)_{fc},(\theta)_{fc}\) & 81.88      \\
\((x,y)_{conv},(w,h)_{strip},(cls)_{fc},(\theta)_{fc}\) & 81.77     \\
\((x,y)_{conv},(w,h)_{conv},(cls)_{fc},(\theta)_{strip}\)& 81.81     \\

\((x,y,w,h)_{strip},(cls)_{fc},(\theta)_{fc}\) & 81.86 \\
\rowcolor[rgb]{0.9,0.9,0.9}
\((x,y,w,h)_{strip},(cls,\theta)_{fc}\) & \textbf{81.96} \\
\((x,y,w,h)_{strip},(cls)_{fc},(\theta)_{strip}\)  & 81.93  \\ \hline
\end{tabular}
\end{table}

\myPara{Effectiveness of the strip head.}
As can be seen in~\tabref{tab:strip_head_apply}, we compare the performance of two different detectors before and after incorporating the strip head. The results demonstrate that our strip head consistently improves the performance of other detectors, confirming its effectiveness and generalizability.
For the ROI Transformer, our method achieves a significant improvement of 5.0 mAP, which is a remarkable increase. Similarly, the other detector Rotated Faster R-CNN also receives performance gains, further validating that our approach can be used for a wide range of detectors and is able to enhance their detection performance.
\begin{table}[t]
\renewcommand\arraystretch{1.2} 
\footnotesize      
\centering
\caption{Effectiveness of strip head on other rotated detectors.}
\label{tab:strip_head_apply}
\setlength{\tabcolsep}{14pt}
\begin{tabular}{l|c|c} 
\hline
Method                              &\textbf{Head}  & \textbf{mAP}~(\%)   \\ 
\hline
\multirow{2}{*}{Faster RCNN-O~\cite{frcnn}}    & Original Head & 76.17 \\
&  \cellcolor[rgb]{0.9,0.9,0.9}Strip Head &\cellcolor[rgb]{0.9,0.9,0.9}77.88  \\
\hline
\multirow{2}{*}{RoI Trans.~\cite{ding_learning_2019}}    & Original Head &74.61  \\
&  \cellcolor[rgb]{0.9,0.9,0.9}Strip Head &\cellcolor[rgb]{0.9,0.9,0.9}79.37  \\ \hline
\end{tabular}
\end{table}

\subsection{Visual Analysis}
\begin{figure}[t]
    \centering
    \includegraphics[width=\linewidth]{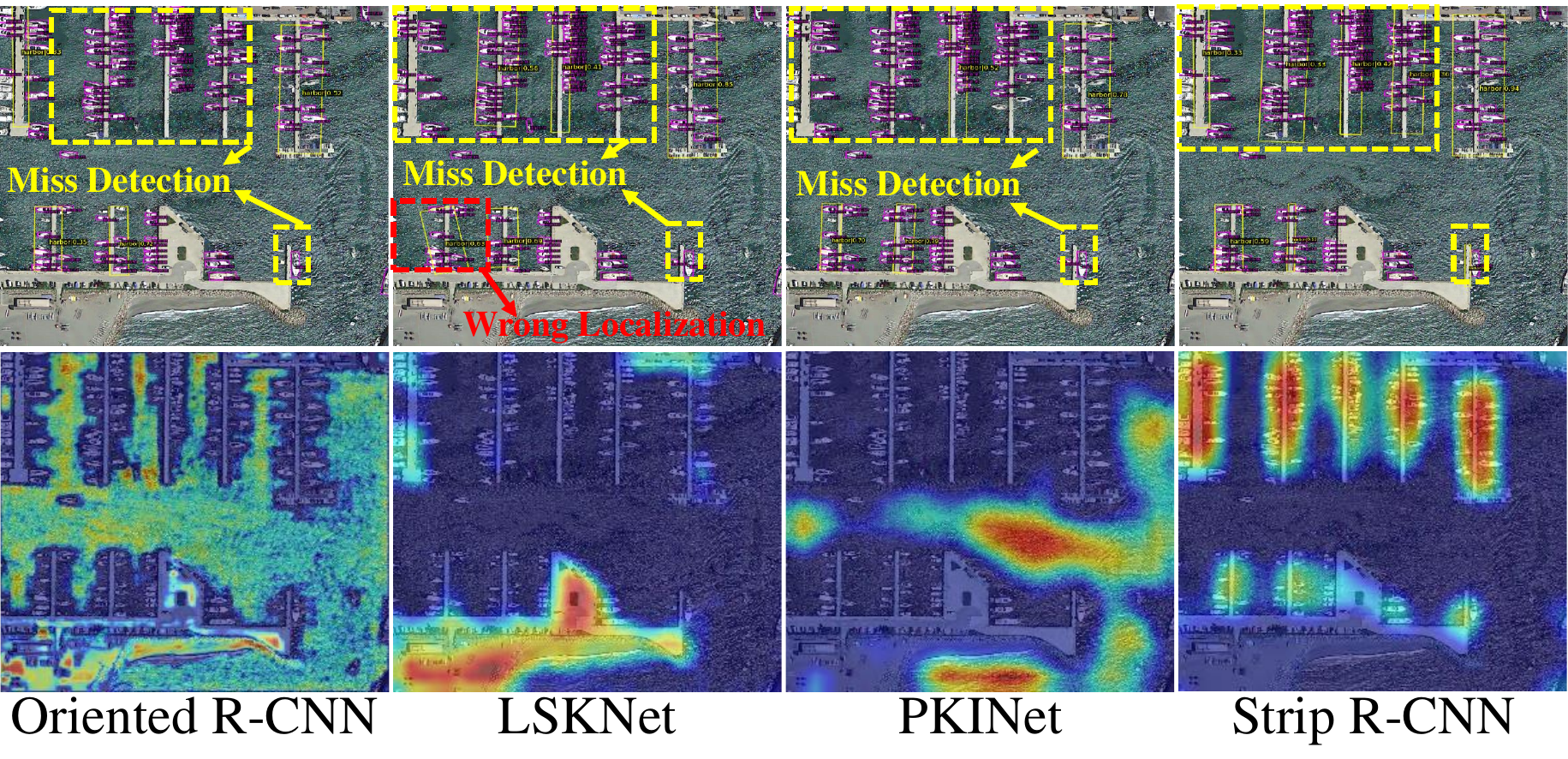}
    \caption{\textbf{Top:} Detection results. Our method \methodname{} can successfully capture the high aspect ratio objects. \textbf{Bottom:} Eigen-CAM visualizations. The Eigen-CAM visualization of our method shows strong activations for high aspect ratio objects, further validating the effectiveness of our approach.}
    \label{fig:result}
\end{figure}

We also present detection results and Eigen-CAM~\cite{eign_cam} visualizations in~\figref{fig:result}. We can see that for high aspect ratio objects, previous methods such as Oriented-RCNN and LSKNet have issues like missed detections and significant localization errors. In contrast, our method can successfully detect the high aspect ratio objects. 
The Eigen-CAM visualizations also show strong activations of our method for these objects, validating the effectiveness of our approach.

Additionally, to further substantiate the superiority of our method in detecting high aspect ratio objects, we conduct additional experiments on the DOTA dataset. The models are trained on the DOTA train set and tested on the val set, following the evaluation method used in~\cite{zheng2024zone}. We assess detection performance across objects with different aspect ratios. As shown in~\figref{fig:DotaStatis}, the results indicate that within the aspect ratio range of 1-5, our method and previous approaches show comparable performance, with only minor differences. However, for objects with larger aspect ratios, our method demonstrates an obvious advantage.

\section{Conclusions}
In this paper, we alleviate the challenge of detecting slender objects in remote sensing scenarios by leveraging large strip convolutions to better extract features and improve localization of such objects. Based on large strip convolutions, we propose the simple yet highly effective \methodname{}. Extensive experiments demonstrate that our method exhibits strong generalization capability and achieves state-of-the-art performance on several remote sensing benchmarks. We hope this research could facilitate the development of object detection in the remote sensing field.

{
\small
\bibliographystyle{ieeenat_fullname}
\bibliography{main}
}

\clearpage
\setcounter{page}{1}
\maketitlesupplementary

\section{Dataset details}
\myPara{DOTA-v1.0~\cite{dota_set}}  is a large-scale dataset for remote sensing detection which contains 2806 images, 188,282 instances, and 15 categories: Plane (PL), Baseball diamond (BD), Bridge (BR), Ground track field (GTF), Small vehicle (SV), Large vehicle (LV), Ship (SH), Tennis court (TC), Basketball court (BC), Storage tank (ST), Soccer-ball field (SBF), Roundabout (RA), Harbor (HA), Swimming pool (SP), and Helicopter (HC). Eeah image is of the size in the range from \(800 \times 800\) to \(20000 \times 20000\) pixels and contains objects exhibiting a wide variety of scales, orientations, and shapes. The proportions of the training set, validation set, and testing set in DOTA-v1.0 are 1/2, 1/6, and 1/3, respectively. For single scale training and testing, we first crop the image into \(1024 \times 1024\) patches with an overlap of 200 following the previous methods~\cite{pu2023adaptive,lsknet,cai2024poly},, we use both the training set and validation set for training, and the testing set for test.

\myPara{DOTA-v1.5~\cite{dota_set}} is a more challenging dataset using the same images as DOTA-v1.0, but the extremely small instances (less than 10 pixels) are also annotated. In this version, a new category, "Container Crane" (CC), has been introduced, and the number of small instances has significantly increased, with a total of 403,318 instances across the dataset. The number of images and dataset splits remains unchanged from DOTA-v1.0.

\myPara{FAIR1M-v1.0~\cite{sun_fair1m_2022}} is a recently published remote sensing dataset that consists of 15,266 high-resolution images and more than 1 million instances. 
All objects in the FAIR1M dataset are annotated with respect to 5 categories and 37 sub-categories by oriented bounding boxes. Each image is of the size in the range from 1000 $\times$ 1000 to 10,000 $\times$ 10,000 pixels and contains objects exhibiting a wide variety of scales, orientations, and shapes. The number of images of the train set and testing set in FAIR1M-1.0 are 16,488 and 8,137, respectively. During training, we apply the same preprocessing methods as DOTA-v1.0 dataset, resulting in multi-scale training and testing sets.
    
\myPara{HRSC2016~\cite{hrsc2016}} is a remote sensing dataset for ship detection that contains 1061 aerial images whose size ranges from \(300 \times 300\) and \(1500 \times 900\). All images are sourced from Google Earth, and the dataset is divided into training, validation, and test sets, with 436 images for training, 181 for validation, and 444 for testing.
    
\myPara{DIOR-R~\cite{DIOR}} dataset extends the DIOR remote sensing dataset by providing oriented bounding box (OBB) annotations. It includes 23,463 images, each with a resolution of 800 × 800 pixels, and a total of 192,518 annotations.

\begin{table}[t]
\renewcommand\arraystretch{1.1} 
\footnotesize      
\centering
\caption{Comparisons with large-kernel-based methods on the DOTA-v1.0~\cite{dota_set} dataset. Params and FLOPs are computed for backbone only. The \backbonename{}-S backbone is pretrained on ImageNet for 300 epochs.}
\label{tab:large_kernel_sup}
\setlength{\tabcolsep}{2.3mm}
\begin{tabular}{l|c|c|c|c} \hline
Method& \textbf{Pre.} & \small\textbf{\#P $\downarrow$} & \small\textbf{FLOPs $\downarrow$} & \textbf{mAP~(\%)} \\
\hline

VAN-B1~\cite{guo_visual_2022} & IN & 13.4M & 52.7G & 81.15 \\
SegNext~\cite{guo_segnext_2022} & IN & 13.1M & 45.0G  & 81.12 \\
ConvNext V2-N~\cite{liu2022convnet} &IN& 15.0M & 51.2G& 80.81 \\

\hline
\rowcolor[rgb]{0.9,0.9,0.9}
$\star$ \methodname{}-S & IN & 13.3M & 52.3G & \textbf{82.28}\\ \hline
\end{tabular}
\end{table}

\section{Implementation details}
In this paper, we adopt the approach outlined in previous works~\cite{lsknet,cai2024poly} to implement multi-scale training and testing across the DOTA, DOTAv1.5, and FAIR1M datasets. Specifically, we rescale each image to three different scales (0.5, 1.0, 1.5) and then crop them into \(1024\times1024\) patches, with a 500-pixel overlap between adjacent patches.
It is important to note that throughout both the pre-training and fine-tuning stages, we did not utilize Exponential Moving Average (EMA) techniques. As a result, the performance metrics reported for LSKNet in the main table are based on results obtained without EMA.

\section{FAIR1M benchmark results}
Fine-grained category result comparisons with state-of-the-art methods on the FAIR1M-v1.0 dataset are given in~\tabref{tab:fair1m_sup}
\begin{table*}[t]
    \renewcommand\arraystretch{1.2} 
    \footnotesize      
    \centering
    \caption{Comparisons With the SOTA Models on the FAIR1M-v1.0 Datasets. \textit{The correspondence between C1-C34 and object categories in the table (in order). C1: Boeing 737, C2: Boeing 747, C3: Boeing 777, C4: Boeing 787, C5: C919, C6: A220, C7: A321, C8: A330, C9: A350, C10: ARJ21, C11: passenger ship, C12: motorboat, C13: fishing boat, C14: tugboat, C15: engineering ship, C16: liquid cargo ship, C17: dry cargo ship, C18: warship, C19: small car, C20: bus, C21: cargo truck, C22: dump truck, C23: van, C24: trailer, C25: tractor, C26: excavator, C27: truck tractor, C28: basketball court, C29: tennis court, C30: football field, C31: baseball field, C32: intersection, C33: roundabout, C34: bridge.}}
    \label{tab:fair1m_sup}
    \setlength{\tabcolsep}{0.8mm}{
    \begin{tabular}{c|c|c|c|c|c|c|c|c|c|c|c|c|c|c|c|c|c|c}
    \hline
    \multirow{2}{*}{\centering \textbf{Method}} & \textbf{C1} & \textbf{C2} & \textbf{C3} & \textbf{C4} & \textbf{C5} & \textbf{C6} & \textbf{C7} & \textbf{C8} & \textbf{C9} & \textbf{C10} & \textbf{C11} & \textbf{C12} & \textbf{C13} & \textbf{C14} & \textbf{C15} & \textbf{C16} & \textbf{C17} & \multirow{2}{*}{\centering \textbf{mAP}} \\
    & \textbf{C18} & \textbf{C19} & \textbf{C20} & \textbf{C21} & \textbf{C22} & \textbf{C23} & \textbf{C24} & \textbf{C25} & \textbf{C26} & \textbf{C27} & \textbf{C28} & \textbf{C29} & \textbf{C30} & \textbf{C31} & \textbf{C32} & \textbf{C33} & \textbf{C34} & (\%)\\
    \hline
     \multirow{2}{*}{\centering Gliding Vertex~\cite{xu_gliding_2021}} & 35.43 & 47.88 & 15.67 & 48.32 & 0.01 & 40.11 & 39.31 & 16.54 & 16.56 & 0.01 & 9.12 & 23.34 & 1.23 & 15.67 & 15.43 & 15.32 & 25.43 & \multirow{2}{*}{\centering 29.92} \\
    & 13.56 & 66.23 & 23.43 & 46.78 & 36.56 & 53.78 & 14.32 & 16.39 & 16.92 & 28.91 & 48.41 & 80.31 & 53.46 & 66.93 & 59.41 & 16.25 & 10.39 & \\
    \hline
    \multirow{2}{*}{\centering RetinaNet~\cite{retinanet}} & 38.46 & 55.36 & 24.57 & 51.84 & 0.81 & 40.5 & 41.06 & 18.02 & 19.94 & 1.7 & 9.57 & 22.54 & 1.33 & 16.37 & 19.11 & 14.26 & 24.7 & \multirow{2}{*}{\centering 30.67} \\
    & 15.37 & 65.2 & 22.42 & 44.17 & 35.57 & 52.44 & 19.17 & 1.28 & 17.03 & 28.98 & 50.58 & 81.09 & 52.5 & 66.76 & 60.13 & 17.41 & 12.58 & \\
    \hline
    \multirow{2}{*}{\centering Cascade-RCNN~\cite{cascade_rcnn}} & 40.42 & 52.86 & 29.07 & 52.47 & 0 & 44.37 & 38.35 & 26.55 & 17.54 & 0 & 12.1 & 28.84 & 0.71 & 15.35 & 18.53 & 14.63 & 25.15 & \multirow{2}{*}{\centering 31.18} \\
    & 14.53 & 68.19 & 28.25 & 48.62 & 40.4 & 58 & 13.66 & 0.91 & 16.45 & 30.27 & 38.81 & 80.29 & 48.21 & 67.9 & 55.67 & 20.35 & 12.62 & \\
    \hline
    \multirow{2}{*}{\centering Faster-RCNN~\cite{frcnn}} & 36.43 & 50.68 & 22.5 & 51.86 & 0.01 & 47.81 & 43.83 & 17.66 & 19.95 & 0.13 & 9.81 & 28.78 & 1.77 & 17.65 & 16.47 & 16.19 & 27.06 & \multirow{2}{*}{\centering 32.12} \\
    & 13.16 & 68.42 & 28.37 & 51.24 & 43.6 & 57.51 & 15.03 & 3.04 & 17.99 & 29.36 & 58.26 & 82.67 & 54.5 & 71.71 & 59.86 & 16.92 & 11.87 & \\
    \hline
    \multirow{2}{*}{\centering Rol Transformer~\cite{ding_learning_2019}} & 39.58 & 73.56 & 18.32 & 56.43 & 0 & 47.67 & 49.91 & 27.64 & 31.79 & 0 & 14.31 & 28.07 & 1.03 & 14.32 & 15.97 & 18.04 & 26.02 & \multirow{2}{*}{\centering 35.29} \\
    & 12.97 & 68.8 & 37.41 & 53.96 & 45.68 & 58.39 & 16.22 & 5.13 & 22.17 & 46.71 & 54.84 & 80.35 & 56.68 & 69.07 & 58.44 & 18.58 & 31.81 & \\
    \hline
    \multirow{2}{*}{\centering OFA-Net~\cite{ming2021oriented}} & 43.20 & 87.58 & 23.58 & 57.78 & 30.71 & 54.31 & 67.83 & 73.65 & 73.56 & 29.25 & 9.90 & 68.15 & 11.70 & 30.37 & 11.83 & 26.80 & 38.68 & \multirow{2}{*}{\centering 43.73} \\
    & 35.42 & 69.62 & 33.67 & 44.69 & 49.23 & 71.61 & 13.45 & 6.06 & 17.36 & 5.35 & 49.20 & 83.75 & 61.29 & 88.77 & 57.97 & 24.68 & 35.82 & \\
    \hline
    \multirow{2}{*}{\centering O-RCNN~\cite{xie_oriented_2021}} & 38.65 & 84.96 & 20.05 & 55.57 & 25.55 & 48.58 & 68.75 & 71.18 & 74.43 & 32.85 & 20.46 & 67.63 & 11.13 & 32.89 & 16.22 & 26.45 & 39.03 & \multirow{2}{*}{\centering 44.30} \\
    & 34.27 & 70.88 & 45.01 & 48.98 & 52.45 & 70.19 & 14.50 & 4.42 & 19.81 & 7.51 & 56.30 & 80.41 & 68.13 & 89.06 & 57.92 & 21.87 & 30.11 & \\
    \hline
    \multirow{2}{*}{\centering CHODNet~\cite{song2022fine}} & 40.23 & 53.39 & 29.65 & 54.04 & 0.02 & 46.26 & 43.12 & 27.61 & 20.89 & 0.01 & 12.34 & 29.34 & 1.71 & 17.77 & 17.72 & 16.78 & 27.51 & \multirow{2}{*}{\centering 32.46} \\
    & 13.68 & 70.12 & 28.38 & 49.11 & 44.02 & 60.78 & 14.48 & 4.96 & 17.60 & 30.09 & 47.91 & 82.11 & 54.10 & 69.97 & 19.57 & 59.91 & 14.38 & \\
    \hline
    \multirow{2}{*}{\centering OBB-ISP~\cite{sun2022ringmo}} & 43.97 & 88.77 & 38.97 & 58.29 & 50.88 & 54.09 & 65.14 & 59.19 & 65.94 & 35.00 & 19.78 & 56.77 & 26.84 & 26.31 & 21.41 & 44.97 & 47.39 & \multirow{2}{*}{\centering 45.91} \\
    & 34.30 & 61.16 & 34.10 & 50.04 & 52.31 & 56.24 & 15.99 & 7.43 & 17.03 & 37.14 & 54.99 & 81.43 & 53.56 & 87.66 & 54.25 & 24.20 & 35.46 & \\
    \hline
    \multirow{2}{*}{\centering LSKNet-S~\cite{lsknet}} & - & - & - & - & - & - & - & - & - & - & - & - & - & - & - & - & - & \multirow{2}{*}{\centering 47.87} \\
    & - & - & - & - & - & - & - & - & - & - & - & - & - & - & - & - & - & \\
    \hline
    \multicolumn{19}{c}{Ours} \\
    \hline
    \multirow{2}{*}{\centering $\star$\methodname{}-S} & 42.17 & 89.07 & 21.52 & 55.24 & 23.03 & 50.97 & 73.58 & 72.96 & 78.29 & 40.81 & 20.78 &  71.52& 16.12 & 42.86 & 17.70 & 28.55 & 40.11 & \multirow{2}{*}{\centering \textbf{48.26}} \\
     & 39.56 & 76.14 & 49.44 & 55.37 & 60.55 & 76.18 & 20.32 & 8.12 & 25.96 & 7.23 & 58.94 & 86.59 & 72.73 & 89.43 &  62.89& 25.57 & 40.58 & \\    \hline
    \end{tabular}}
\end{table*}

\section{DOTA benchmark results}
Fine-grained category result comparisons with state-of-the-art methods on the DOTA-v1.0 and DOTA-v1.5 dataset are given in~\tabref{tab:dota15ss_sup} and ~\tabref{dota1.0_sup}

\begin{table*}[t]
  \renewcommand\arraystretch{1.1} 
  \setlength{\tabcolsep}{1.8pt}
  \footnotesize      
  \centering
  \caption{Comparisons with SOTA methods on the DOTA-v1.0 dataset with single-scale and multi-scale training and testing. 
  The \backbonename{}-S backbone is pretrained on ImageNet for 300 epochs. $^\dagger$: Model ensemble as in MoCAE~\cite{oksuz2023mocae}.
  }\label{tab:dota1.0_sup}
\begin{tabular}{l|c|cccccccccccccccccc} \hline
  Method &\textbf{Pre.} & \textbf{mAP $\uparrow$} & \textbf{\#P $\downarrow$} & \textbf{FLOPs $\downarrow$} & PL & BD & BR & GTF & SV & LV & SH & TC & BC & ST & SBF & RA & HA & SP & HC \\ \hline
  \multicolumn{20}{c}{\textit{Single-Scale}} \\ \hline
EMO2-DETR~\cite{hu2023emo2} & IN & 70.91 & 74.3M  & 304G 
& 87.99 & 79.46 & 45.74 & 66.64 & 78.90 & 73.90 & 73.30 & 90.40 & 80.55 & 85.89 & 55.19 & 63.62 & 51.83 & 70.15 & 60.04  \\
CenterMap~\cite{wang_learning_2021} & IN   & 71.59   & 41.1M & 198G  &
 89.02 & 80.56 & 49.41 & 61.98 & 77.99 &  74.19 & 83.74 & 89.44 & 78.01 & 83.52 & 47.64 & 65.93 & 63.68 & 67.07 & 61.59 \\
AO2-DETR~\cite{dai_ao2-detr_2022} & IN & 72.15 &  74.3M & 304G 
& 86.01 & 75.92 & 46.02 & 66.65 & 79.70 & 79.93 & 89.17 & 90.44 & 81.19 & 76.00 & 56.91 & 62.45 & 64.22 & 65.80 & 58.96 \\
SCRDet~\cite{yang_scrdet_2019}    & IN  & 72.61  & 41.9M & - &
89.98 & 80.65 & 52.09 & 68.36 & 68.36 & 60.32 & 72.41 & 90.85 & 87.94 & 86.86 & 65.02 & 66.68 & 66.25 & 68.24 & 65.21 \\
R3Det~\cite{yang_r3det_nodate}    & IN    & 73.70 & 41.9M & 336G 
& 89.5 & 81.2 & 50.5 & 66.1 & 70.9 & 78.7 & 78.2 & 90.8 & 85.3 & 84.2 
& 61.8 & 63.8 & 68.2 & 69.8 & 67.2 \\
Rol Trans.~\cite{ding_learning_2019}  & IN   & 74.05 & 55.1M &  200G & 
 89.01 & 77.48 & 51.64 & 72.07 & 74.43 & 77.55 & 87.76 & 90.81 & 79.71 & 85.27 & 58.36 & 64.11 & 76.50 & 71.99 & 54.06 \\
S$^2$ANet~\cite{han_align_2020}     & IN  & 74.12  & 38.5M & - 
&  89.11 & 82.84 & 48.37 & 71.11 & 78.11 & 78.39 & 87.25 & 90.83 & 84.90 & 85.64 & 60.36 & 62.60 & 65.26 &  69.13 & 57.94 \\
SASM~\cite{SASM}   & IN   & 74.92 & 36.6M & - 
& 86.42 & 78.97 & 52.47 & 69.84 & 77.30 & 75.99 & 86.72 & 90.89 & 82.63 & 85.66 & 60.13 & 68.25 & 73.98 & 72.22 & 62.37 \\
G.V.~\cite{xu_gliding_2021}    & IN  & 75.02   &  41.1M & 198G & 
89.64 & 85.00 & 52.26 & 77.34 & 73.01 & 73.14 & 86.82 & 90.74 & 79.02 & 86.81 & 59.55 & 70.91 & 72.94 & 70.86 & 57.32 \\
O-RCNN~\cite{xie_oriented_2021}    & IN   & 75.87   & 41.1M &  199G & 
89.46 & 82.12 & 54.78 & 70.86 & 78.93 & 83.00 & 88.20 & 90.90 & 87.50 & 84.68 & 63.97 & 67.69 & 74.94 & 68.84 & 52.28 \\
ReDet~\cite{han_redet_2021}   & IN   & 76.25  & 31.6M &  -  & 
88.79 & 82.64 & 53.97 & 74.00 & 78.13 & 84.06 & 88.04 & 90.89 & 87.78 & 85.75 & 61.76 & 60.39 & 75.96 & 68.07 & 63.59  \\
R3Det-GWD~\cite{yang_rethinking_2021}   & IN    & 76.34 & 41.9M & 336G & 
88.82 & 82.94 & 55.63 & 72.75 & 78.52 &  83.10 & 87.46 & 90.21 & 86.36 & 85.44 & 64.70 & 61.41 & 73.46 & 76.94 & 57.38 \\
COBB~\cite{xiao2024theoretically} & IN & 76.52 & 41.9M & - &
- & - & - & - & - & - & - & - & - & - & - & - & - & - & - \\
R3Det-KLD~\cite{yang_learning_2021}  & IN   & 77.36   & 41.9M & 336G &
88.90 & 84.17 & 55.80 & 69.35 & 78.72 & 84.08 & 87.00 & 89.75 & 84.32 & 85.73 & 64.74 & 61.80 & 76.62 & 78.49 & 70.89 \\
ARC~\cite{yang_kfiou_2022}  & IN   & 77.35   & 74.4M & 217G  & 
89.40 & 82.48 & 55.33 & 73.88 & 79.37 & 84.05 & 88.06 & 90.90 & 86.44 & 84.83 & 63.63 & 70.32 & 74.29 & 71.91 & 65.43 \\
LSKNet-S~\cite{lsknet} & IN &   77.49   & 31.0M  & 161G  & 
89.66 & 85.52 & 57.72 & 75.70 & 74.95 & 78.69 & 88.24 & 90.88 & 86.79 & 86.38 & 66.92 & 63.77 & 77.77 & 74.47 & 64.82  \\
PKINet-S~\cite{cai2024poly} & IN & 78.39 & 30.8M & 190G  &
89.72 & 84.20 & 55.81 & 77.63 & 80.25 & 84.45 & 88.12 & 90.88 & 87.57 & 86.07 & 66.86 & 70.23 & 77.47 & 73.62 & 62.94 \\
RTMDet-R~\cite{lyu_rtmdet_2022} & IN & 78.85 & 52.3M & 205G & 89.43 & 84.21 & 55.20 & 75.06 & 80.81 & 84.53 & 88.97 & 90.90 & 87.38 & 87.25 & 63.09 & 67.87 & 78.09 & 80.78 & 69.13 \\
\hline
\rowcolor[rgb]{0.9,0.9,0.9}$\star$ \methodname{}-S  & IN &  \textbf{80.06}  & \textbf{30.5M}  & \textbf{159G}  & 
88.91 &  86.38 & 57.44 &  76.37 &  79.73 &  84.38 &  88.25 &  90.86 &  86.71 &  87.45 &  69.89 &  66.82 &  79.25 &  82.91 &  75.58 \\
\hline
\multicolumn{20}{c}{\textit{Multi-Scale}}  \\ 
\hline
G.V.~\cite{xu_gliding_2021}    & IN  & 75.02   &  41.1M & 198G  & 89.64 & 85.00 & 52.26 & 77.34 & 73.01 & 73.14 & 86.82 & 90.74 & 79.02 & 86.81 & 59.55 & 70.91 & 72.94 & 70.86 & 57.32  \\
R3Det~\cite{yang_r3det_nodate}    & IN    & 76.47 & 41.9M & 336G & 89.80 & 83.77 & 48.11 & 66.77 & 78.76 & 83.27 & 87.84 & 90.82 & 85.38 & 85.51 & 65.57 & 62.68 & 67.53 & 78.56 & 72.62  \\
CSL~\cite{yang_arbitrary-oriented_2020}   & IN    & 76.17   & 37.4M & 236G  & 90.25 & 85.53 & 54.64 & 75.31 & 70.44 & 73.51 & 77.62 & 90.84 & 86.15 & 86.69 & 69.60 & 68.04 & 73.83 & 71.10 & 68.93  \\
CFA~\cite{guo_beyond_2021}   & IN   & 76.67  & - & - & 89.08 & 83.20 & 54.37 & 66.87 & 81.23 & 80.96 & 87.17 & 90.21 & 84.32 & 86.09 & 52.34 & 69.94 & 75.52 & 80.76 & 67.96  \\
DAFNet~\cite{DAFNe}   & IN   & 76.95 &-&-  & 89.40 & 86.27 & 53.70 & 60.51 & 82.04 & 81.17 & 88.66 & 90.37 & 83.81 & 87.27 & 53.93 & 69.38 & 75.61 & 81.26 & 70.86  \\
S$^2$ANet~\cite{han_align_2020}     & IN  & 79.42  & - & - & 88.89 & 83.60 & 57.74 & 81.95 & 79.94 & 83.19 & 89.11 & 90.78 & 84.87 & 87.81 & 70.30 & 68.25 & 78.30 & 77.01 & 69.58  \\
R3Det-GWD~\cite{yang_rethinking_2021}   & IN    & 80.23 & 41.9M & 336G & 89.66 & 84.99 & 59.26 & 82.19 & 78.97 & 84.83 & 87.70 & 90.21 & 86.54 & 86.85 & 73.47 & 67.77 & 76.92 & 79.22 & 74.92  \\
DODet~\cite{DODet}   & IN  & 80.62   &  -&  - & 89.96 & 85.52 & 58.01 & 81.22 & 78.71 & 85.46 & 88.59 & {90.89} & 87.12 & 87.80 & 70.50 & 71.54 & 82.06 & 77.43 & 74.47  \\
AOPG~\cite{DIOR} & IN   & 80.66 &-& - & 89.88 & 85.57 & 60.90 & 81.51 & 78.70 & 85.29 & 88.85 & {90.89} & 87.60 & 87.65 & 71.66 & 68.69 & 82.31 & 77.32 & 73.10  \\
R3Det-KLD~\cite{yang_learning_2021}  & IN   & 80.63 & 41.9M & 336G & 89.92 & 85.13 & 59.19 & 81.33 & 78.82 & 84.38 & 87.50 & 89.80 & 87.33 & 87.00 & 72.57 & 71.35 & 77.12 & 79.34 & 78.68 \\
KFloU~\cite{yang_kfiou_2022}  & IN   & 80.93   & 58.8M & 206G  & 89.44 & 84.41 & 62.22 & 82.51 & 80.10 & 86.07 & 88.68 & 90.90 & 87.32 & 88.38 & 72.80  & 71.95 & 78.96 & 74.95 & 75.27  \\ 
PKINet-S~\cite{cai2024poly} & IN & 81.06 & 30.8M & 190G  &
89.02 & 86.73 & 58.95 & 81.20 & 80.41 & 84.94 & 88.10 & 90.88 & 86.60 & 87.28 & 67.10 & 74.81 & 78.18 & 81.91 & 70.62 \\
RTMDet-R~\cite{lyu_rtmdet_2022}   & CO    & 81.33  & 52.3M & 205G & 88.01 & 86.17 & 58.54 & 82.44 & 81.30 & 84.82 & 88.71 & {90.89} & 88.77 & 87.37 & 71.96 & 71.18 & 81.23 & 81.40 & 77.13  \\ 
RVSA~\cite{wang_advancing_2022}  & MA & 81.24   & 114.4M & 414G & 88.97 & 85.76 & 61.46 & 81.27 & 79.98 & 85.31 & 88.30  & 90.84 & 85.06 & 87.50 & 66.77 & 73.11 & 84.75 & 81.88 & 77.58  \\ 
LSKNet-S~\cite{lsknet} & IN &   81.64   & 31.0M  & 161G  &  89.57   & 86.34 &  63.13 & 83.67  & 82.20  &  86.10 & 88.66 & {90.89}  & 88.41  & 87.42 & 71.72 & 69.58 & 78.88  & 81.77 & 76.52  \\
\hline
\rowcolor[rgb]{0.9,0.9,0.9}$\star$ \methodname{}-T  & IN &    81.40    & 20.5M &  123G  &   89.14    &  84.90     & 61.78      & 83.50     & 81.54      & 85.87      & 88.64      & {90.89}      & 88.02      & 87.31      & 71.55      & 70.74      & 78.66      & 79.81      & 78.16       \\
\rowcolor[rgb]{0.9,0.9,0.9}$\star$ \methodname{}-S  & IN &  \textbf{82.28}  & \textbf{30.5M}  & \textbf{159G}  &  89.17   &  85.57 &   62.40 & 83.71  & 81.93  &  86.58 & 88.84 & 90.86 & 87.97  & 87.91 & 72.07 & 71.88 & 79.25  & 82.45 & 82.82  \\ 
\rowcolor[rgb]{0.9,0.9,0.9}$\star$ \methodname{}-S$^\dagger$  & IN &  \textcolor{RoyalBlue}{\textbf{82.75}}   & 30.5M  & 159G  &  88.99   &  86.56 &   61.35 & 83.94  & 81.70  &  85.16 & 88.57 & 90.88  & 88.62  & 87.36 & 75.13 & 74.34 & 84.58  & 81.49 & 82.56 \\ \hline
\end{tabular}
\end{table*}

\begin{table*}[t]
\renewcommand\arraystretch{1.2} 
\footnotesize      
\centering
\caption{Comparison with SOTA methods on the \textbf{DOTA-v1.5} dataset with single-scale training and testing.}
\label{tab:dota15ss_sup}

\setlength{\tabcolsep}{2.6pt}
\begin{tabular}{l|c|cccccccccccccccccc} 
\hline
Method                              &\textbf{Pre.}  & PL    & BD    & BR    & GTF   & SV    & LV    & SH    & TC    & BC    & ST    & SBF   & RA    & HA    & SP    & HC  & CC & \textbf{mAP(\%) $\uparrow$}    \\ 
\hline

RetinaNet-O~\cite{retinanet} & IN & 71.43 & 77.64 & 42.12 & 64.65 & 44.53 & 56.79 & 73.31 & 90.84 & 76.02 & 59.96 & 46.95 & 69.24 & 59.65 & 64.52 & 48.06 & 0.83 & 59.16 \\
FR-O~\cite{frcnn} & IN & 71.89 & 74.47 & 44.45 & 59.87 & 51.28 & 68.98 & 79.37 & 90.78 & 77.38 & 67.50 & 47.75 & 69.72 & 61.22 & 65.28 & 60.47 & 1.54 & 62.00 \\
Mask R-CNN~\cite{maskrcnn}  & IN &  76.84 & 73.51 & 49.90 & 57.80 & 51.31 & 71.34  & 79.75 & 90.46 & 74.21 & 66.07 & 46.21 & 70.61 & 63.07 & 64.46 & 57.81 & 9.42 & 62.67 \\
HTC~\cite{Chen2019HybridTC} & IN &  77.80 & 73.67 & 51.40 & 63.99 & 51.54 & 73.31 & 80.31 & 90.48 & 75.12  & 67.34  & 48.51 & 70.63 & 64.84 & 64.48 & 55.87 & 5.15 & 63.40 \\
ReDet~\cite{han_redet_2021} & IN & 79.20 & 82.81 & 51.92 & 71.41 & 52.38 & 75.73 & 80.92 & 90.83 & 75.81 & 68.64  & 49.29 & 72.03 & 73.36 & 70.55 & 63.33 & 11.53 & 66.86 \\
LSKNet-S~\cite{lsknet} & IN & 72.05 & 84.94 & 55.41 & 74.93 & 52.42 & 77.45 & 81.17 & 90.85 & 79.44 & 69.00 & 62.10 & 73.72 & 77.49 & 75.29 & 55.81 & 42.19 & 70.26 \\
PKINet-S~\cite{cai2024poly} & IN &  80.31 & 85.00 & 55.61 & 74.38 & 52.41 & 76.85 & 88.38 & 90.87 & 79.04 & 68.78 & 67.47 & 72.45 & 76.24 & 74.53 & 64.07 & 37.13 & 71.47 \\
\hline
\rowcolor[rgb]{0.9,0.9,0.9}$\star$ \methodname{}-S  & IN &  80.04 & 83.26 & 54.40 & 75.38 & 52.46 & 81.44 & 88.53 & 90.83 & 84.80 & 69.65 & 65.93 & 73.28 & 74.61 & 74.04 & 69.70 & 38.98 & \textbf{72.27}\\ 
\hline
\end{tabular}
\end{table*}

\begin{figure*}[t]
    \centering
    \includegraphics[width=0.8\linewidth]{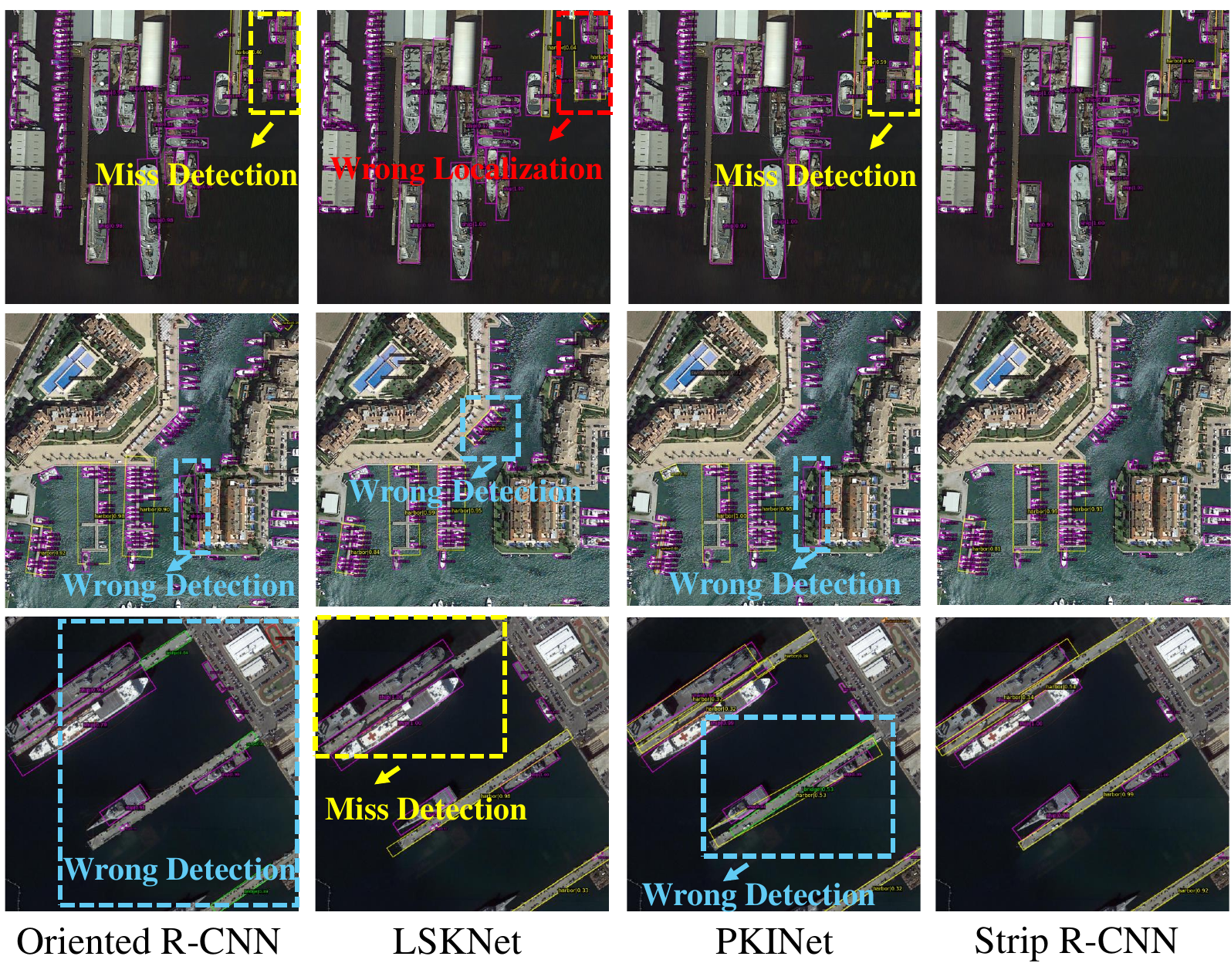}
    \caption{Detection results. Our method \methodname{}-S can successfully capture the high aspect ratio objects which validating the effectiveness of our approach.}
    \label{fig:detection_results_sup}
\end{figure*}

\begin{table*}[htbp]
\renewcommand\arraystretch{1.2} 
\footnotesize      
\centering
\caption{Effectiveness of \backbonename{}-S backbone on other remote sensing object detection frameworks. The \backbonename{}-S backbone is pretrained on ImageNet~\cite{imagenet} for 100 epochs.}
\label{tab:stripnet_backbone_sup}
\setlength{\tabcolsep}{2.7pt}
\begin{tabular}{l|c|cccccccccccccccc} 
\hline
Method                              &\textbf{Backbone}  & PL    & BD    & BR    & GTF   & SV    & LV    & SH    & TC    & BC    & ST    & SBF   & RA    & HA    & SP    & HC  & \textbf{mAP~(\%)}   \\ 
\hline
\multirow{2}{*}{RoI Trans.~\cite{ding_learning_2019}}    & ResNet-50~\cite{he2016deep} & 
88.65 & 82.60 & 52.53 & 70.87 & 77.93 & 76.67 & 86.87 & 90.71 & 83.83 & 82.51 & 53.95 & 67.61 & 74.67 &  68.75 & 61.03 & 74.61 \\
&  \cellcolor[rgb]{0.9,0.9,0.9}\backbonename{}-S & \cellcolor[rgb]{0.9,0.9,0.9}89.10 &\cellcolor[rgb]{0.9,0.9,0.9}85.45 & \cellcolor[rgb]{0.9,0.9,0.9}63.61 &\cellcolor[rgb]{0.9,0.9,0.9}82.84 & \cellcolor[rgb]{0.9,0.9,0.9}80.95 & \cellcolor[rgb]{0.9,0.9,0.9}85.84 & \cellcolor[rgb]{0.9,0.9,0.9}88.68 & \cellcolor[rgb]{0.9,0.9,0.9}90.84 & \cellcolor[rgb]{0.9,0.9,0.9}88.35 & \cellcolor[rgb]{0.9,0.9,0.9}87.92 & \cellcolor[rgb]{0.9,0.9,0.9}72.12 & \cellcolor[rgb]{0.9,0.9,0.9}71.37 & \cellcolor[rgb]{0.9,0.9,0.9}79.73 & \cellcolor[rgb]{0.9,0.9,0.9}83.32 & \cellcolor[rgb]{0.9,0.9,0.9}75.77 &\cellcolor[rgb]{0.9,0.9,0.9}\textbf{81.72}  \\

\hline
\hline
\multirow{2}{*}{O-RCNN~\cite{xie_oriented_2021}}    & ResNet-50~\cite{he2016deep} & 89.84 & 85.43 & 61.09 & 79.82 & 79.71 & 85.35 & 88.82 & 90.88 & 86.68 & 87.73 & 72.21 & 70.80 & 82.42 & 78.18 & 74.11& 80.87 \\
& \cellcolor[rgb]{0.9,0.9,0.9}\backbonename{}-S & \cellcolor[rgb]{0.9,0.9,0.9}89.46 & \cellcolor[rgb]{0.9,0.9,0.9}84.40 & \cellcolor[rgb]{0.9,0.9,0.9}63.04 & \cellcolor[rgb]{0.9,0.9,0.9}82.89 & \cellcolor[rgb]{0.9,0.9,0.9}81.84 & \cellcolor[rgb]{0.9,0.9,0.9}86.16 & \cellcolor[rgb]{0.9,0.9,0.9}88.66 & \cellcolor[rgb]{0.9,0.9,0.9}90.88 & \cellcolor[rgb]{0.9,0.9,0.9}88.12 & \cellcolor[rgb]{0.9,0.9,0.9}86.97 & \cellcolor[rgb]{0.9,0.9,0.9}74.27 & \cellcolor[rgb]{0.9,0.9,0.9}73.69& \cellcolor[rgb]{0.9,0.9,0.9}79.05 & \cellcolor[rgb]{0.9,0.9,0.9}81.14 & \cellcolor[rgb]{0.9,0.9,0.9}76.44 & \cellcolor[rgb]{0.9,0.9,0.9}\textbf{81.75}  \\
\hline
\hline
\multirow{2}{*}{S$^2$ANet~\cite{han_align_2020}}    & ResNet-50~\cite{he2016deep} & 88.89 & 83.60 & 57.74&  81.95 & 79.94 & 83.19 & 89.11 & 90.78 & 84.87 & 87.81 & 70.30 & 68.25 & 78.30 & 77.01 & 69.58 & 79.42 \\
& \cellcolor[rgb]{0.9,0.9,0.9}\backbonename{}-S & \cellcolor[rgb]{0.9,0.9,0.9}89.31 & \cellcolor[rgb]{0.9,0.9,0.9}84.94 & \cellcolor[rgb]{0.9,0.9,0.9}61.01 & \cellcolor[rgb]{0.9,0.9,0.9}81.87 & \cellcolor[rgb]{0.9,0.9,0.9}81.34 & \cellcolor[rgb]{0.9,0.9,0.9}85.10 & \cellcolor[rgb]{0.9,0.9,0.9}88.69 & \cellcolor[rgb]{0.9,0.9,0.9}90.76 & \cellcolor[rgb]{0.9,0.9,0.9}86.89 & \cellcolor[rgb]{0.9,0.9,0.9}87.49 & \cellcolor[rgb]{0.9,0.9,0.9}67.94 & \cellcolor[rgb]{0.9,0.9,0.9}68.74& \cellcolor[rgb]{0.9,0.9,0.9}77.96 & \cellcolor[rgb]{0.9,0.9,0.9}78.82 & \cellcolor[rgb]{0.9,0.9,0.9}73.90 & \cellcolor[rgb]{0.9,0.9,0.9}\textbf{80.32}  \\

\hline
\multirow{2}{*}{R3Det~\cite{yang_r3det_nodate}} & ResNet-50~\cite{he2016deep}& 89.80 & 83.77 & 48.11 & 66.77 & 78.76 & 83.27 & 87.84 & 90.82 & 85.38 & 85.51 & 65.57 & 62.68 & 67.53 & 78.56 & 72.62 & 76.47\\
& \cellcolor[rgb]{0.9,0.9,0.9}\backbonename{}-S & \cellcolor[rgb]{0.9,0.9,0.9}89.56 & \cellcolor[rgb]{0.9,0.9,0.9}83.47 & \cellcolor[rgb]{0.9,0.9,0.9}56.71 & \cellcolor[rgb]{0.9,0.9,0.9}81.28 & \cellcolor[rgb]{0.9,0.9,0.9}79.80& \cellcolor[rgb]{0.9,0.9,0.9}83.87 & \cellcolor[rgb]{0.9,0.9,0.9}88.63 & \cellcolor[rgb]{0.9,0.9,0.9}90.87 & \cellcolor[rgb]{0.9,0.9,0.9}86.16 & \cellcolor[rgb]{0.9,0.9,0.9}86.93 & \cellcolor[rgb]{0.9,0.9,0.9}66.82 & \cellcolor[rgb]{0.9,0.9,0.9}69.16& \cellcolor[rgb]{0.9,0.9,0.9}75.64 & \cellcolor[rgb]{0.9,0.9,0.9}74.42 & \cellcolor[rgb]{0.9,0.9,0.9}71.88 & \cellcolor[rgb]{0.9,0.9,0.9}\textbf{79.01}  \\
\hline
\end{tabular}
\end{table*}

\begin{table*}[htbp]
\renewcommand\arraystretch{1.2} 
\footnotesize      
\centering
\caption{Effectiveness of strip head on other remote sensing object detectors.}
\label{tab:strip_head_sup}
\setlength{\tabcolsep}{2.7pt}
\begin{tabular}{l|c|cccccccccccccccc} 
\hline
Method                              &\textbf{Head}  & PL    & BD    & BR    & GTF   & SV    & LV    & SH    & TC    & BC    & ST    & SBF   & RA    & HA    & SP    & HC  & \textbf{mAP~(\%)}   \\ 
\hline
\multirow{2}{*}{LSKNet-S~\cite{lsknet}}    & Original Head & 
89.66 & 85.52 & 57.72 & 75.70 & 74.95 & 78.69 & 88.24 & 90.88 & 86.79 & 86.38 & 66.92 & 63.77 & 77.77 & 74.47 & 64.82 & 77.49 \\
&  \cellcolor[rgb]{0.9,0.9,0.9}Strip Head & \cellcolor[rgb]{0.9,0.9,0.9}89.71 &\cellcolor[rgb]{0.9,0.9,0.9}83.50 & \cellcolor[rgb]{0.9,0.9,0.9}56.18 &\cellcolor[rgb]{0.9,0.9,0.9}78.86 & \cellcolor[rgb]{0.9,0.9,0.9}79.83 & \cellcolor[rgb]{0.9,0.9,0.9}84.66 & \cellcolor[rgb]{0.9,0.9,0.9}88.16 & \cellcolor[rgb]{0.9,0.9,0.9}90.86 & \cellcolor[rgb]{0.9,0.9,0.9}88.30 & \cellcolor[rgb]{0.9,0.9,0.9}86.11 & \cellcolor[rgb]{0.9,0.9,0.9}67.02 & \cellcolor[rgb]{0.9,0.9,0.9}66.27 & \cellcolor[rgb]{0.9,0.9,0.9}76.19 & \cellcolor[rgb]{0.9,0.9,0.9}71.79 & \cellcolor[rgb]{0.9,0.9,0.9}63.39&\cellcolor[rgb]{0.9,0.9,0.9}\textbf{78.05}  \\

\hline
\hline
\multirow{2}{*}{O-RCNN~\cite{xie_oriented_2021}}    & Original Head & 89.46 & 82.12 & 54.78 & 70.86 & 78.93 & 83.00 & 88.20 & 90.90 & 87.50 & 84.68 & 63.97 & 67.69 & 74.94 & 68.84 & 52.28 
& 75.81 \\
& \cellcolor[rgb]{0.9,0.9,0.9}Strip Head & \cellcolor[rgb]{0.9,0.9,0.9}89.48 & \cellcolor[rgb]{0.9,0.9,0.9}82.33 & \cellcolor[rgb]{0.9,0.9,0.9}53.95 & \cellcolor[rgb]{0.9,0.9,0.9}72.83 & \cellcolor[rgb]{0.9,0.9,0.9}78.60 & \cellcolor[rgb]{0.9,0.9,0.9}77.97 & \cellcolor[rgb]{0.9,0.9,0.9}88.12 & \cellcolor[rgb]{0.9,0.9,0.9}90.89 & \cellcolor[rgb]{0.9,0.9,0.9}85.84 & \cellcolor[rgb]{0.9,0.9,0.9}84.76 & \cellcolor[rgb]{0.9,0.9,0.9}64.42 & \cellcolor[rgb]{0.9,0.9,0.9}67.55 & \cellcolor[rgb]{0.9,0.9,0.9}74.57 & \cellcolor[rgb]{0.9,0.9,0.9}70.31 & \cellcolor[rgb]{0.9,0.9,0.9}60.05 & \cellcolor[rgb]{0.9,0.9,0.9}\textbf{76.11}  \\
\hline
\end{tabular}
\end{table*}

\section{More results of the effectiveness of the strip head}
More result comparisons of two different detectors before and after incorporating the strip head are shown in~\tabref{tab:strip_head_sup}. The results is obtained on DOTA-v1.0 dataset with single-scale training and testing. 
As can be seen in~\tabref{tab:strip_head_sup}, we compare the performance of two different detectors before and after incorporating the strip head. The results demonstrate that our strip head consistently improves the performance of other detectors, confirming its effectiveness and generalizability.
For LSKNet~\cite{lsknet}, our method achieves a improvement of 0.56\%. Similarly, Oriented R-CNN~\cite{xie_oriented_2021} also receives a 0.3\% performance gain, further validating that our approach is compatible with a wide range of detectors and is able to reliably enhance detection performance.

\section{The effectiveness of the \backbonename{} backbone}
To validate the generality and effectiveness of our proposed \backbonename{} backbone, we conduct experiments using various remote sensing detection frameworks. These include two-stage frameworks such as O-RCNN~\cite{xie_oriented_2021} and RoI Transformer~\cite{ding_learning_2019}, as well as one-stage frameworks like S$^2$ANet~\cite{han_align_2020} and R3Det~\cite{yang_r3det_nodate}.
Result comparisons of two different detectors before and after incorporating the \backbonename{}-S are shown in~\tabref{tab:stripnet_backbone_sup}. The results is obtained on DOTA-v1.0 dataset with multi-scale training and testing. 
As can be seen in~\tabref{tab:stripnet_backbone_sup}, we compare the performance of two different detectors before and after incorporating the \backbonename{}-S backbone. The results demonstrate that our \backbonename{}-S backbone consistently improves the performance of other detectors, confirming its effectiveness and generalizability.
For RoI Transformer~\cite{ding_learning_2019}, our method achieves a improvement of 7.11\%. Similarly, Oriented R-CNN~\cite{xie_oriented_2021} also receives a 0.88\% performance gain, S$^2$ANet and R3Det obtains a improvement of  0.9\% and 2.54\% respectively, further validating that our approach is compatible with a wide range of detectors and is able to reliably enhance detection performance.

\section{Comparisons with other large kernel networks}
We also compare our \methodname{}-S with some popular high-performance backbone models which using large kernels in\tabref{tab:large_kernel_sup}.
In the three models under discussion, all have adopted large convolutional kernels. Specifically, VAN~\cite{guo_visual_2022} utilizes dilated convolutions to expand the receptive field. While this technique effectively increases the model's receptive feild, it can also lead to the neglect of certain critical information, which is particularly disadvantageous for densely distributed objects in remote sensing scenes. SegNext~\cite{guo_segnext_2022} introduces a multi-branch large convolutional kernel structure, designed to capture richer feature information through multiple pathways. However, this architecture suffers from issues of feature confusion and feature redundancy, which may impact the model's performance.
Although these three methods validate the effectiveness of large convolutional kernels, they fail to adequately address the unique characteristics of data specific to remote sensing scenarios. This limitation restricts their generalization capabilities in such domains. In response to this challenge, our proposed method not only thoroughly analyzes the features of remote sensing data but also specifically optimizes for objects with high aspect ratios. Consequently, our solution demonstrates superior performance in remote sensing applications, offering new insights and directions for addressing challenges in this field.

\section{More detection results}
We present additional detection results in~\figref{fig:detection_results_sup}. As can be seen in~\figref{fig:detection_results_sup}, previous methods, including Oriented R-CNN, LSKNet, and PKINet, frequently struggle with detecting high aspect ratio objects, often resulting in missed or incorrect detections. In contrast, our approach, \methodname{}, effectively detects these challenging objects.

\end{document}